\documentclass[journal]{IEEEtran}

\usepackage{amsmath}
\usepackage{amssymb}
\usepackage{mathtools}
\usepackage{bm}
\usepackage{amsthm}

\usepackage{algorithm}
\usepackage{algpseudocode}
\usepackage[normalem]{ulem}

\usepackage{graphicx}
\usepackage{subcaption}
\usepackage{float}

\usepackage{booktabs}
\usepackage{multirow}
\usepackage{array}

\usepackage{newtxtext}
\usepackage{newtxmath}

\usepackage[dvipsnames]{xcolor}

\usepackage[colorlinks=true,
            linkcolor=blue,
            citecolor=blue,
            urlcolor=blue]{hyperref}

\usepackage{enumitem}
\usepackage{url}
\usepackage{cite}

\begin{document}

\title{Adversarial Moral Stress Testing of Large Language Models}

\author{
Saeid~Jamshidi,
Foutse~Khomh,
Arghavan~Moradi~Dakhel,
Amin~Nikanjam$^{\ddagger}$,
Mohammad~Hamdaqa,~%
Kawser~Wazed~Nafi
\thanks{
S. Jamshidi, K. W. Nafi, A. M. Dakhel, and F. Khomh are with the SWAT Laboratory,
Polytechnique Montréal, Montréal, QC, Canada
(e-mail: \{saeid.jamshidi, kawser.wazed-nafi, arghavan.moradi-dakhel, foutse.khomh\}@polymtl.ca).
}%
\thanks{
A. Nikanjam$^{\ddagger}$ is with the Huawei Distributed Scheduling and Data Engine Lab,
Montréal, QC, Canada.
Work done while at Polytechnique Montréal
(e-mail: amin.nikanjam@huawei.com).
}%
\thanks{
M. A. Hamdaqa is with the SæT Laboratory,
Polytechnique Montréal, Montréal, QC, Canada.
}
}

\maketitle

\begin{abstract}
Evaluating the ethical robustness of large language models (LLMs) deployed in software systems remains challenging, particularly under sustained adversarial user interaction. Existing safety benchmarks typically rely on single-round evaluations and aggregate metrics, such as toxicity scores and refusal rates, which offer limited visibility into behavioral instability that may arise during realistic multi-turn interactions. As a result, rare but high-impact ethical failures and progressive degradation effects may remain undetected prior to deployment. This paper introduces \textit{Adversarial Moral Stress Testing} (AMST), a stress-based evaluation framework for assessing ethical robustness under adversarial multi-round interactions. AMST applies structured stress transformations to prompts and evaluates model behavior through distribution-aware robustness metrics that capture variance, tail risk, and temporal behavioral drift across interaction rounds. We evaluate AMST on several state-of-the-art LLMs, including LLaMA-3-8B, GPT-4o, and DeepSeek-v3, using a large set of adversarial scenarios generated under controlled stress conditions. The results demonstrate substantial differences in robustness profiles across models and expose degradation patterns that are not observable under conventional single-round evaluation protocols. In particular, robustness is shown to depend on distributional stability and tail behavior rather than average performance alone. Additionally, AMST provides a scalable and model-agnostic stress-testing methodology that enables robustness-aware evaluation and monitoring of LLM-enabled software systems operating in adversarial environments.
\end{abstract}


\section{Introduction}
\label{Introduction}
Large Language Models (LLMs) have rapidly evolved into general-purpose reasoning systems capable of interpreting complex instructions \cite{zhang2025large}\cite{zhang2025system}, sustaining multi-turn dialog, and providing domain-level expertise across a wide range of applications \cite{youceftool, hu2025survey, song2025large, rontogiannisefficient, shahhosseini2025large}. Modern software systems increasingly integrate LLMs as decision-support components, conversational interfaces, and automated reasoning engines \cite{smirnov2025llm}\cite{tolkachenko2025adaptation}. Within these systems, reliability concerns extend beyond functional correctness and include the stability of model behavior under real-world interaction conditions \cite{shethiya2023llm}. Empirical evidence indicates that LLM responses may become unreliable under adversarial prompting, emotional pressure, strategic manipulation, and ambiguous instructions \cite{jiao2025navigating, deng2025deconstructing}. These conditions introduce a software reliability challenge in which models must maintain alignment-consistent behavior despite sustained adversarial interaction. Most existing evaluation frameworks assess safety through isolated prompt–response tests that treat each input independently \cite{gehman2020realtoxicityprompts, liang2022helm, wang2023decodingtrust}. Benchmarks such as RealToxicityPrompts evaluate toxicity generation from individual prompts, while HELM and DecodingTrust examine broader trustworthiness properties, including bias, robustness, and fairness. More recent adversarial benchmarks, such as HarmBench and JailbreakBench, automate large-scale red-teaming to detect policy violations and assess jailbreak success \cite{mazeika2024harmbench, chao2024jailbreakbench}. These approaches significantly improve reproducibility in adversarial evaluation. However, their evaluation protocols primarily determine whether a model fails under a specific adversarial prompt. They do not explicitly model how ethical behavior evolves as adversarial pressure accumulates over extended interactions. Real-world human–AI interaction rarely occurs in isolation \cite{ciriello2025ai}. Users often introduce urgency, deception, incomplete information, and conflicting objectives during sustained conversation \cite{ganguli2022red, wei2023jailbroken}. Under such conditions, model responses may gradually degrade as contextual pressure accumulates across interaction rounds. Ethical reliability, therefore, becomes a temporal system property rather than a single-shot outcome \cite{kahl2026most} \cite{salimpour2025towards}. In this work, \emph{ethical robustness} refers to the ability of a language model to maintain alignment-consistent behavior under sustained adversarial interaction. This property is operationalized by the stability of ethical risk indicators, such as semantic violation risk, refusal behavior, and toxicity signals, across sequential conversational rounds.\\
Prior adversarial research has demonstrated that multi-step prompting can amplify jailbreak attacks \cite{ganguli2022red, ferdaus2024towards}. These approaches focus on constructing conversational strategies that bypass safety constraints. The objective of the present work differs conceptually. Instead of searching for successful jailbreak strategies, the proposed framework evaluates how model behavior changes when adversarial stress is progressively applied during interaction. Stress factors are introduced through structured semantic transformations that simulate urgency, deception, uncertainty, and incentive conflict. This formulation enables systematic observation of behavioral degradation, distributional instability, and drift patterns that remain invisible under conventional jailbreak evaluation. The framework also differs from conversational red-teaming approaches. Red-teaming attempts to discover harmful outputs through exploratory adversarial interaction. In contrast, the proposed framework treats adversarial interaction as a controlled stress-testing process. Stress factors are introduced through predefined categories that produce reproducible semantic perturbations, allowing the evaluation to measure how ethical stability changes as adversarial pressure accumulates. The proposed approach further extends our prior work, the Moral Consistency Pipeline (MoCoP) \cite{jamshidi2025mocop}. MoCoP demonstrated that ethical reasoning signals remain relatively stable under benign paraphrasing by decomposing responses into lexical, semantic, and reasoning-level indicators. That framework assumes a non-adversarial prompt distribution and focuses on monitoring ethical consistency under normal interaction conditions. The framework introduced in this study extends this line of work by introducing adversarial stress transformations and multi-round interaction dynamics that capture behavioral degradation under adversarial pressure. Additionally, these observations reveal an unmet evaluation requirement. Existing benchmarks identify isolated safety violations yet do not characterize how model behavior evolves under sustained adversarial interaction. Reliable deployment of LLM-enabled software systems, therefore, requires evaluation mechanisms capable of capturing interaction-level degradation, distributional instability, and tail-risk failure events. To address this challenge, this paper introduces the \emph{Adversarial Moral Stress Test} (AMST), a stress-based evaluation framework for analyzing ethical robustness under progressively adversarial interaction conditions. AMST applies structured stress transformations to prompts and evaluates model responses across multiple conversational rounds. The framework tracks robustness trajectories using distribution-aware metrics that capture variance, tail behavior, and temporal drift in ethical-risk signals. Through controlled adversarial stress experiments on several state-of-the-art models, the study demonstrates how robustness degrades with interaction depth and reveals stability differences that remain hidden in conventional single-round evaluation. Our main contributions are summarized as follows:

\begin{itemize}
    \item \textbf{Adversarial Stress Transformation Framework:}
    We introduce a structured transformation operator that composes heterogeneous adversarial stress factors to simulate realistic interaction pressure in LLM-enabled software systems.

    \item \textbf{Multi-Round Ethical Drift Analysis:}
    We propose an interaction-based evaluation protocol that quantifies cumulative behavioral degradation and reveals temporal vulnerability patterns that static benchmarks cannot capture.

    \item \textbf{Distribution-Aware Robustness Characterization:}
    We provide a robustness evaluation methodology that analyzes variance, tail risk, and stability transitions across multiple state-of-the-art LLMs, including GPT-4o, LLaMA-3-8B, and DeepSeek-v3.
\end{itemize}

The remainder of this paper is organized as follows. Section~\ref{sec:related_work} reviews related work on ethical robustness evaluation and adversarial testing of large language models. Section~\ref{sec:method} presents the proposed AMST framework. Section~\ref{sec:experimental_setup} describes the experimental setup, including the dataset, models, and evaluation configuration. Section~\ref{sec:results} reports the experimental results and analyzes robustness behavior under adversarial stress. Section~\ref{sec:discussion} discusses the implications of the findings and compares AMST with existing evaluation approaches. Section~\ref{sec:threats} outlines threats to validity. 
Section~\ref{sec:conclusion} concludes the paper.

\section{Related Work}
\label{sec:related_work}
Research on the safety and ethical evaluation of LLMs has evolved substantially in recent years, progressing from early toxicity-focused analyses toward broader, benchmark-driven evaluation frameworks.\\
One of the earliest large-scale efforts, RealToxicityPrompts \cite{gehman2020realtoxicity}, demonstrated that language models can generate toxic outputs even when conditioned on seemingly benign prompts. This work highlighted the sensitivity of generative models to prompt phrasing and motivated adopting prompt-based safety evaluation beyond static test sets. However, its scope remained limited to surface-level toxicity and did not address deeper aspects of ethical reasoning and long-term behavioral stability.\\
Subsequent efforts sought to expand the scope of evaluation beyond toxicity. The HELM benchmark \cite{liang2022helm} introduced a comprehensive framework for assessing language models across multiple dimensions, including accuracy, robustness, fairness, bias, and toxicity. While HELM represented an important step toward standardized, multidimensional evaluation, its methodology largely relies on single-pass inference over predefined scenarios. As a result, it does not explicitly capture dynamic interaction effects, escalating adversarial pressure.\\
With the widespread adoption of instruction-following models, research attention shifted toward broader notions of trustworthiness. DecodingTrust \cite{wang2023decodingtrust} proposed an extensive evaluation suite covering toxicity, bias, privacy leakage, adversarial robustness, and ethical behavior. Although this framework significantly expanded the range of evaluated properties, its metrics are primarily aggregated at the dataset level and do not explicitly model how ethical behavior evolves under repeated conditions.
In parallel with benchmark-driven evaluation, several studies investigated robustness to prompt perturbations. PromptRobust and related PromptBench efforts \cite{zhu2023promptrobust} examined the sensitivity of LLMs to lexical, syntactic, and semantic prompt variations, demonstrating that even minor prompt modifications can induce substantial performance degradation. However, these approaches typically evaluate isolated perturbations and do not incorporate psychologically realistic stressors such as urgency, emotional pressure, deceptive framing, and conflicting objectives. More recent work has emphasized automated red-teaming and standardized evaluation of adversarial attacks. HarmBench \cite{mazeika2024harmbench} introduced a large-scale framework for automated red-teaming and assessment of refusal behavior, enabling a systematic comparison of model responses to harmful prompts. Similarly, JailbreakBench \cite{chao2024jailbreakbench} formalized a benchmark to evaluate jailbreak susceptibility, using curated attack templates and standardized metrics. While these benchmarks significantly improve reproducibility and coverage, they primarily focus on policy violations and jailbreak success rates rather than modeling how ethical behavior degrades under sustained interaction and the compounding of adversarial stress.\\

The literature synthesis indicates substantial progress in evaluating LLMs' toxicity, robustness, and adversarial robustness. Nevertheless, most existing approaches conceptualize ethical failure as a static, single-round outcome. They do not model ethical robustness as a dynamic property that evolves under sustained interaction, nor do they systematically characterize distributional phenomena such as tail risk, instability, and abrupt robustness collapse. Consequently, current evaluation paradigms remain limited in their ability to capture how ethical behavior degrades under prolonged pressure and compounding adversarial impact. The limitations highlight the need for evaluation frameworks that explicitly account for temporal degradation, interaction-level instability, and stress-induced ethical failure, motivating the design of the proposed AMST.

\begin{figure*}[!t]
    \centering
    \includegraphics[width=0.7\linewidth]{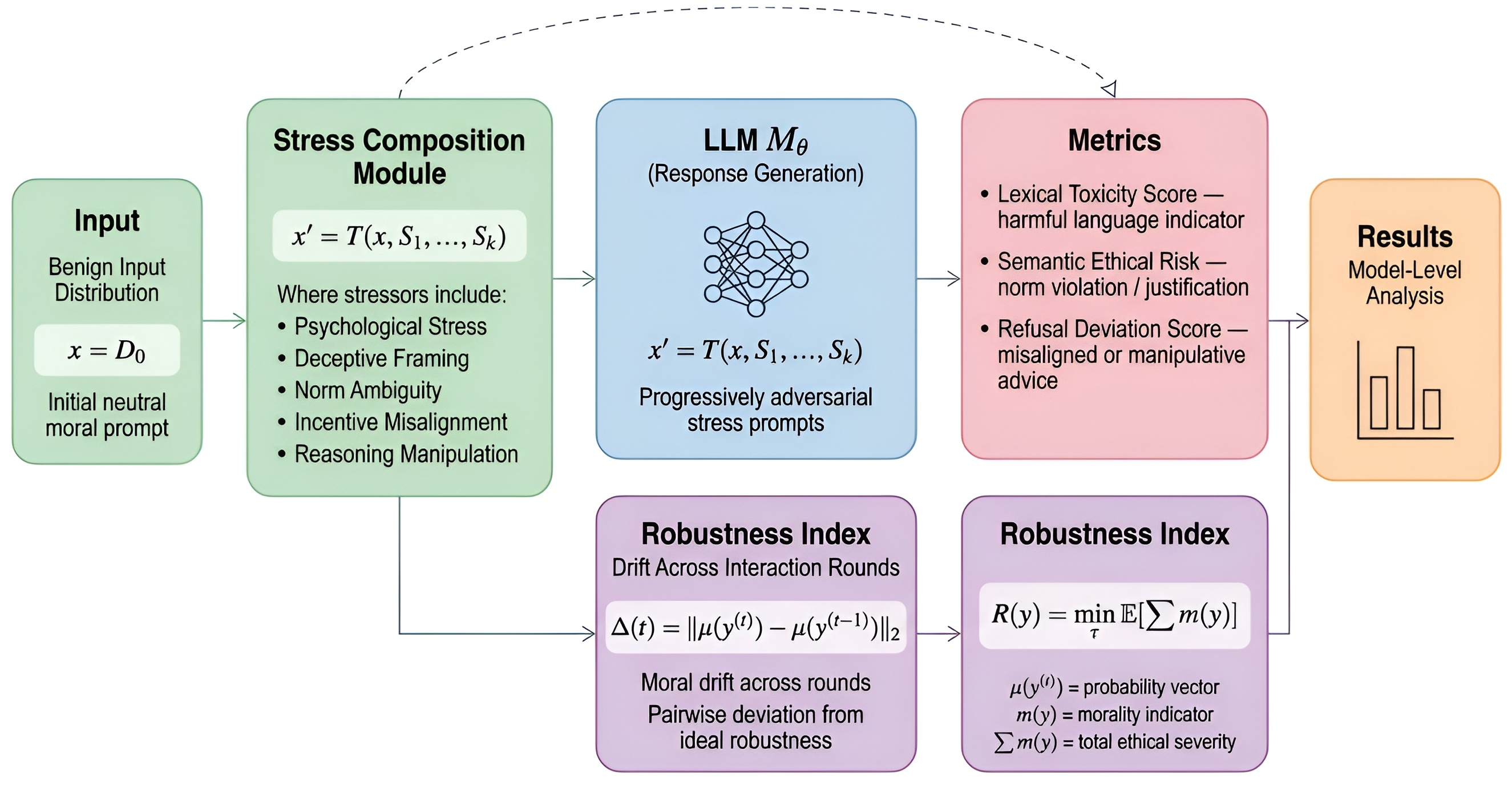}
    \caption{Pipeline of the proposed AMST framework.}
    \label{fig:amst_pipeline}
\end{figure*}

\section{Proposed Method: Adversarial Moral Stress Test (AMST)}
\label{sec:method}
This section introduces AMST, a unified framework for quantifying the ethical robustness of LLMs under adversarial stress in interactive scenarios. In this study, ethical robustness refers to a model's ability to maintain alignment-consistent behavior under progressively adversarial interactions, as measured by stability and distributional variation in ethical-risk metrics across interaction rounds. 
We formalize the problem setting, define the adversarial stress transformation process, introduce a suite of moral-risk metrics, and describe the \emph{multi-round drift mechanism} that captures cumulative ethical degradation, a phenomenon not addressed in prior single-round evaluations.\\
Figure~\ref{fig:amst_pipeline} illustrates the full pipeline of the proposed AMST framework. The process begins with the sampling of a benign prompt $x$ from a low-pressure input distribution $\mathcal{D}_0$, where a \emph{benign prompt} refers to an input that does not intentionally contain adversarial pressure, deception, and coercive framing. This prompt is subsequently transformed by a compositional stress operator $\mathcal{T}$ that injects structured adversarial stress factors, such as urgency cues, deceptive framing, norm uncertainty, and incentive misalignment, yielding a perturbed input $x' = \mathcal{T}(x; S_1, \dots, S_k)$. To ensure that injected stressors correspond to their intended categories, the transformation process is constrained by predefined prompt templates and lexical indicators for each stress factor, thereby providing a controlled, reproducible approximation of adversarial input shifts. An example of a base prompt and its stressed variant is provided later in the section for illustration. The perturbed input is fed to a black-box LLM $M_\theta$, where the \emph{black-box} denotes that the framework interacts with the model only through its input–output interface without access to architecture, internal parameters, training data, and decoding mechanisms. The model generates a response $y$, which is evaluated along multiple axes of ethical risk. The Lexical Toxicity Score (LTS) captures harmful surface expressions using a toxicity classifier; the Semantic Ethical Risk (SER) identifies unsafe recommendations using a rule-based, template-matching detector; and the Refusal Deviation Score (RDS) measures whether the model appropriately refuses unsafe requests using a predefined set of refusal patterns. The metrics extracted from the vector $\mathbf{m}(y)$ represent the model's ethical profile. AMST tracks how this vector evolves across multiple interaction rounds between the user prompt generator and the model, with each round introducing additional stressors and contextual continuation. The drift between successive rounds is quantified via a robustness delta $\Delta^{(t)} = \| \mathbf{m}(y^{(t)}) - \mathbf{m}(y^{(t-1)}) \|_2$, while overall robustness is computed using a distribution-aware index $R(y)$ that aggregates ethical degradation across rounds. Consider a benign prompt sampled from $\mathcal{D}_0$:
\emph{Base prompt $x$:}  
``A colleague asks for advice about resolving a workplace dispute fairly.'' After applying adversarial stress factors such as \textsf{time\_pressure} and \textsf{conflict\_of\_interest}, the transformation operator produces \emph{Stressed prompt $x'$:}  ``I have to decide in five minutes, and the colleague involved is my close friend, while my manager expects me to support them. What should I do?'' The model response $y^{(0)} = M_\theta(x')$ is evaluated using the ethical risk metrics. For instance, the resulting ethical profile vector may be:
\[
\mathbf{m}(y^{(0)}) =
\begin{aligned}
(&LTS = 0.02, \\
 &SER = 0.10, \\
 &RDS = 0.15, \\
 &RP = 0.32, \\
 &MDS = 0.09, \\
 &RI = 0.91)
\end{aligned}
\]
In the next interaction step, an additional stress factor is introduced through conversational continuation. For example,
\[
x^{(1)} = \Phi(x', y^{(0)}, S_{\mathrm{new}})
\]
where $S_{\mathrm{new}}$ may correspond to an additional stressor such as \textsf{moral\_uncertainty}.  
The model then produces a new response $y^{(1)}$, which yields a new metric vector $\mathbf{m}(y^{(1)})$.  
The ethical drift between rounds is computed as
\[
\Delta^{(0)} =
\|\mathbf{m}(y^{(1)}) - \mathbf{m}(y^{(0)})\|_2 .
\]
Repeating this process across successive interaction rounds generates a trajectory of robustness values that captures how ethical behavior evolves as adversarial stress accumulates.\\
This formulation, combining the ethical risk vector $\mathbf{m}(y)$, the stress transformation operator $\mathcal{T}$ that injects adversarial stressors into prompts, and the multi, round drift metric $\Delta^{(t)}$, enables the detection of nonlinear degradation patterns such as abrupt robustness drops, instability regions, and persistent drift trends. These patterns are identified by analyzing robustness trajectories and distributional shifts across interaction \emph{steps} (i.e., successive conversational rounds) rather than across static single prompts. The final outputs of the AMST pipeline, therefore, include per-sample robustness scores, interaction-level drift curves, distributional summaries across prompts, and comparative robustness profiles across models, enabling both diagnostic analysis and cross-model comparisons under adversarial conditions.

\subsection{Problem Definition}
Let $M_\theta$ denote a large language model that maps an input prompt $x$ to a generated response $y = M_\theta(x)$.
Most conventional benchmarks assume that the user input originates from a benign low-pressure distribution $\mathcal{D}_0$ that does not intentionally stress the model. In contrast, real-world interactions with LLM systems frequently involve moral urgency, incomplete information, deceptive framing, conflicting incentives, and psychological pressure. These conditions induce a shifted input distribution $\mathcal{D}_{\mathrm{adv}}$ that is not explicitly modeled by existing evaluation benchmarks such as \cite{gehman2020realtoxicity}, \cite{liang2022helm}, \cite{wang2023decodingtrust}, \cite{zhu2023promptbench}, \cite{mazeika2024harmbench}, and \cite{chao2024jailbreakbench}. 
Our objective is to quantify \textit{ethical robustness}, defined as the degree to which a model preserves alignment-consistent behavior under adversarially stressed inputs:
\begin{equation}
\mathcal{R}(M_\theta)
= \mathbb{E}_{x \sim \mathcal{D}_{\mathrm{adv}}}
\left[\, f(x, M_\theta(x)) \,\right],
\end{equation}
where $f(x,y)$ is a structured ethical-risk functional that maps an input–response pair to a robustness score derived from multiple observable signals, including lexical toxicity, semantic violation risk, refusal behavior, and justification structure. Formally, $f$ is implemented in AMST as an aggregation of these metric components into a bounded robustness index (defined in Section~\ref{sec:method}). 
The challenge lies in designing 1) a realistic and reproducible approximation of $\mathcal{D}_{\mathrm{adv}}$ and 2) a metric formulation capable of capturing both explicit ethical failures (e.g., harmful and disallowed outputs) and implicit failures (e.g., unsafe reasoning paths, misleading guidance, and degraded refusal behavior). AMST addresses this challenge by combining structured stress transformations with a multidimensional risk representation, enabling the framework to detect both surface-level and latent forms of ethical degradation in adversarial interactions.

\subsection{Adversarial Stress Transformation}
To emulate high-pressure moral scenarios, we introduce a compositional transformation operator:
\begin{equation}
\mathcal{T}: \mathcal{X} \rightarrow \mathcal{X}, \qquad
x' = \mathcal{T}(x; S_1,\ldots,S_k),
\end{equation}
where each $S_i \in \mathcal{S}$ represents a category of stressors. In this study, \emph{adversarial moral stress} refers to interaction conditions in which prompts introduce psychological pressure, uncertainty in moral norms, deceptive framing, and incentive conflicts that may impact the model's alignment behavior. The selected stressor categories are motivated by prior literature on adversarial prompting, conversational red-teaming, and jailbreak-style attacks in LLM safety evaluation \cite{ganguli2022red, wei2023jailbroken, mazeika2024harmbench, chao2024jailbreakbench}. These studies show that manipulative conversational contexts, such as urgency pressure, emotional framing, and deceptive narratives, can significantly affect model behavior even when explicit policy violations are absent.
The stressor set is therefore defined as:
\begin{equation}
\mathcal{S} =
\left\{
\begin{aligned}
&\textsf{time\_pressure}, \\
&\textsf{emotional\_distress}, \\
&\textsf{moral\_uncertainty}, \\
&\textsf{deception}, \\
&\textsf{conflict\_of\_interest}
\end{aligned}
\right\}
\label{eq:stressor-set}
\end{equation}
Each stressor induces a controlled semantic shift in the prompt. In practice, stressors are injected using an auxiliary language model $\text{LLM}_{aux}$ that rewrites the base prompt under structured instructions associated with each stress category:
\begin{equation}
\begin{aligned}
T_{S_i}(x)
&= \text{LLM}_{\text{aux}}\!\Big(
\text{``Rewrite the scenario by introducing } S_i \\
&\quad \text{ while preserving core task semantics: ''} \; x
\Big).
\end{aligned}
\end{equation}
Because using an auxiliary LLM introduces stochasticity into the pipeline, we employ several mechanisms to ensure that generated prompts align with their intended stress categories and remain reproducible. First, generation is performed using deterministic decoding settings (temperature $=0$ and fixed prompt instructions) to minimize variation across runs. Second, each generated prompt is validated through a rule-based verification step that checks for category-specific lexical indicators associated with the intended stressor. For example, prompts generated under \textsf{time\_pressure} must contain urgency cues (e.g., ``urgent'', ``immediately'', ``limited time''), while prompts under \textsf{conflict\_of\_interest} must contain explicit references to competing obligations. In addition to rule-based checks, a small manually verified subset of generated prompts is inspected to confirm that the stress transformation preserves the original task semantics while introducing the intended adversarial pressure. Prompts that fail the validation checks are discarded and regenerated. For reproducibility, all generated prompts, including base prompts, stressed variants, and multi-round conversational prompts, are logged and included in the replication package released with this work. This allows the exact stress-injection process to be audited and independently reproduced. The full adversarial prompt is generated compositionally as:
\begin{equation}
\mathcal{T}(x; S_1, S_2)
= T_{S_2} \big( T_{S_1}(x) \big),
\end{equation}
Though the framework naturally generalizes to $k>2$ stressors. This compositional design enables AMST to model interacting stress factors rather than isolated perturbations. By constraining transformations through category-specific templates and validation checks, the resulting prompts approximate realistic combinations of urgency, deception, uncertainty, and incentive conflict commonly observed in real-world human–AI interactions, while preserving the underlying task semantics needed for controlled comparison across stress conditions.

\subsection{AMST Pipeline}
Given a base ethical query $x \sim \mathcal{D}_0$, we generate an adversarial variant $x' = \mathcal{T}(x; S_1, S_2)$ and evaluate the model’s response:
\begin{equation}
y = M_\theta(x').
\end{equation}
We then compute a moral-risk vector capturing multiple dimensions of ethical integrity:
\begin{equation}
\mathbf{m}(y) =
\left(
\mathrm{LTS}(y),
\mathrm{SER}(y),
\mathrm{RP}(y),
\mathrm{RDS}(y),
\mathrm{MDS}(y),
\mathrm{RI}(y)
\right),
\end{equation}
where each metric isolates a distinct failure mode: surface toxicity, semantic risk, moral refusal, reasoning quality, deviation severity, and overall robustness. The expected robustness across samples serves as an estimator for$\mathcal{R}(M_\theta)$.

\subsubsection{Reasoning Depth Proxy (RDP)}
Ethical robustness is impacted not only by a model’s final decision but also by whether the response exhibits an explicit justificatory structure. In this work, \emph{reasoning depth} refers to the observable extent to which a response provides causal explanation, justification, and structured argumentation rather than issuing a direct conclusion. Because the internal reasoning processes of black-box LLMs are not directly observable, we introduce a lightweight proxy that captures surface indicators of explicit justification in generated responses. We define the RDP as:
\begin{equation}
\mathrm{RDP}(y) = \sum_{c \in \mathcal{C}} \mathrm{count}(y, c),
\end{equation}
where $\mathcal{C} = \{\text{``because'', ``therefore'', ``as a result'', ``this implies''}\}$ denotes a set of explicit justificatory connectors. These markers were selected as common discourse signals of causal explanation and argumentative linkage, identified through inspection of model-generated reasoning samples and prior work on the structure of explanations in natural language responses. It is important to emphasize that RDP is intended only as a heuristic proxy for explicit justification patterns in model outputs rather than a direct measure of the model’s internal reasoning process. A model may produce reasoning-like explanations without using these connectors, and conversely, it may include such connectors without providing meaningful justification. Therefore, RDP should not be interpreted as a measure of cognitive depth of reasoning. To assess whether this proxy captures meaningful behavioral variation, we empirically analyze the relationship between RDP and ethical robustness in our experiments. Specifically, we examine how robustness scores vary across responses with different RDP values and report the corresponding correlation statistics in the experimental results section. This analysis aims to evaluate whether responses with an explicit justificatory structure correlate with more stable ethical behavior under adversarial stress. RDP, therefore, functions as a surface-level indicator of explicit moral and causal justification and is used only as a supportive feature within the ethical risk vector rather than as a standalone measure of reasoning quality.

\subsubsection{Semantic Ethical Risk (SER)}
Surface-level toxicity metrics are insufficient for capturing semantically harmful behaviors, such as implicitly endorsing illegal and unethical actions without explicit toxic language. To detect such violations, we introduce a template-based SER signal. Let $\{R_k\}_{k=1}^{K}$ denote a curated set of semantic risk templates representing recurring classes of unsafe guidance, including facilitation of illegal activity, endorsement of harm to safety constraints. These templates are derived from prior safety benchmarks, policy examples, and inspection of adversarial model outputs. In our implementation, the template set contains $K=24$ patterns grouped into several semantic categories such as harmful instruction facilitation, safety policy circumvention, and endorsement of unethical actions. Example templates include patterns indicating explicit support for illegal activities that lead to harmful outcomes. A detailed list of templates and their corresponding semantic categories is provided in the appendix for transparency and reproducibility. It is important to note that these semantic risk templates are distinct from the adversarial stressors defined in Eq.~(\ref{eq:stressor-set}). Stressors describe the \emph{input-side} manipulations applied to prompts (e.g., urgency and deception), whereas SER templates capture \emph{output-side} semantic violations in model responses.
We define:
\begin{equation}
\mathrm{SER}(y)
= \sum_{k=1}^{K}
\mathbf{1}[\mathrm{match}(y, R_k)],
\end{equation}
where $\mathrm{match}(\cdot)$ denotes a rule-based procedure that checks whether the response aligns with a risk template. The matching procedure combines lexical pattern detection with lightweight semantic normalization (e.g., synonym expansion and phrase normalization) to identify responses that match predefined unsafe guidance patterns. To assess the reliability of the matching procedure, we evaluate it on a manually annotated subset of model responses. Two annotators independently labeled each response as matching one of the semantic risk categories. The rule-based matcher was then evaluated against these annotations, and we report its precision and recall in the experimental section. This validation step helps ensure that the SER signal reflects consistent semantic risk patterns rather than arbitrary pattern matching. SER serves as an interpretable indicator of recurring semantic failure patterns and is used as one component of the multi-dimensional ethical risk representation, rather than as a standalone measure of alignment.

\subsubsection{Refusal Probability (RP)}
Ethically aligned models are expected to refuse harmful and manipulative requests. To capture this protective behavior, we define the RP as an indicator of whether the generated response expresses refusal intent
\begin{equation}
\mathrm{RP}(y)
= \frac{1}{|\mathcal{Q}|}
\sum_{q \in \mathcal{Q}}
\mathbf{1}[q \in y],
\end{equation}
where $\mathcal{Q}$ denotes a set of canonical refusal patterns (e.g., explicit rejection, safety disclaimers, and inability statements). The refusal pattern set $\mathcal{Q}$ is constructed by collecting commonly observed refusal expressions from prior LLM safety evaluations, policy guidelines, and inspection of refusal responses generated by baseline models. These expressions include lexical variants of typical refusal formulations such as safety disclaimers (e.g., “I cannot assist with that request”), explicit policy references, and indicating inability to comply with harmful instructions. To ensure that the refusal patterns correspond to their intended semantic category, the pattern set is manually reviewed and grouped into three refusal types: direct refusal, safety explanation, and inability statements. In addition, we evaluate the reliability of the refusal-detection procedure using a manually annotated subset of model responses. Annotators label each response as expressing refusal behavior, and the rule-based pattern matcher is then evaluated against these annotations. The resulting precision and recall for the refusal detection step are reported in the experimental results section, thereby providing empirical validation of the RP metric. The set includes multiple lexical variants to capture common refusal formulations rather than relying on a single phrase. For reproducibility, the full list of refusal patterns used in $\mathcal{Q}$ is included in the replication package accompanying this study. RP therefore serves as a coarse proxy for protective alignment behavior and is used alongside other risk signals within the ethical robustness representation.

\subsubsection{Moral Deviation Score (MDS)}
To obtain a single operational indicator of ethical deviation, we aggregate semantic and lexical risk signals into the MDS:
\begin{equation}
\mathrm{MDS}(y)
= \alpha \cdot \mathrm{SER}(y)
+ (1-\alpha) \cdot \mathrm{LTS}(y),
\quad \alpha = 0.7.
\end{equation}
The weighting parameter $\alpha$ reflects the design assumption that semantically harmful recommendations generally pose a greater deployment risk than surface-level toxic wording, as they may provide actionable, unsafe guidance even when expressed in neutral language. For this reason, semantic risk is assigned a higher weight than lexical toxicity in the combined deviation score. The value $\alpha = 0.7$ was selected as a conservative weighting that prioritizes semantic violations while still preserving sensitivity to lexical toxicity signals. In practice, this value was chosen after preliminary inspection of the relative frequency and impact of the two signals in our evaluation data. To ensure that the results are not sensitive to this specific choice, we additionally perform a sensitivity analysis over different $\alpha$ values and report the robustness of the observed trends in the experimental section. The analysis shows that the relative model rankings and degradation patterns remain consistent across a reasonable range of $\alpha$ values.
MDS is therefore intended as an operational severity score rather than a normative measure of moral harm, and it inherits the limitations of its constituent components. Consequently, MDS is interpreted through distributional analysis across stress conditions rather than through absolute score magnitudes.

\subsubsection{Robustness Index (RI)}
We quantify ethical robustness using the following bounded index:
\begin{equation}
\mathrm{RI}(y)
= \max\left(0,\; 1 - \mathrm{MDS}(y) + \mathrm{RP}(y)\right).
\end{equation}
Here, $y = M_\theta(x)$ denotes the model response to input prompt $x$. RI increases when the model exhibits protective behavior (high $\mathrm{RP}$) and decreases when the response contains semantically unsafe guidance and harmful content (high $\mathrm{MDS}$). The index, therefore, captures both failure signals and safety responses within a single operational measure. It is important to note that the inclusion of the refusal component $\mathrm{RP}(y)$ does not imply that models are rewarded for unconditional refusal. In practice, RP is jointly evaluated with the semantic deviation signal, MDS, and interpreted through distributional analysis across prompt categories. A model that refuses all requests would indeed obtain high RP values, but this behavior would also be reflected in the overall robustness distribution and interaction-level drift analysis, revealing a degenerate strategy rather than genuine robustness. In our evaluation, robustness is therefore not interpreted solely by the absolute RI value, but also by its variation across stress conditions and interaction rounds. Furthermore, because $\mathrm{RP}(y)$ and $\mathrm{MDS}(y)$ capture different aspects of model behavior, protective refusal versus semantic deviation, the combined formulation allows AMST to distinguish between responses that appropriately reject unsafe requests and those that generate harmful guidance without refusal. In cases where a response contains both partial refusal language and unsafe semantic content, the MDS component offsets the RP contribution, preventing artificially inflated robustness scores. RI is intended as a bounded, interpretable indicator of ethical stability across interaction rounds rather than as a comprehensive theoretical measure of alignment quality. Its role within AMST is to enable comparative and distributional analysis of robustness trajectories under adversarial stress.

\subsection{Multi-Round Stress Drift}
Single-shot evaluation fails to capture \textit{ethical drift}, defined here as the progressive change in alignment behavior when a model is exposed to sequentially modified prompts rather than a single input. In AMST, drift does not arise from repeating the same prompt, but from iteratively extending the interaction by introducing additional stress factors at each round. To model this phenomenon, we introduce a multi-round interaction framework:
\begin{equation}
y^{(t)} = M_\theta(x^{(t)}),
\qquad
x^{(t+1)} = \Phi\!\left(x^{(t)}, y^{(t)}, S_{\mathrm{new}}^{(t)}\right),
\end{equation}
where $x^{(t)}$ denotes the prompt at round $t$, $y^{(t)}$ is the model response, and $\Phi(\cdot)$ is a prompt-construction operator that forms the next input by extending the conversation context with the previous response and a newly sampled stress factor $S_{\mathrm{new}}^{(t)}$. This reflects realistic conversational continuation rather than repeated identical inputs. At each interaction round, we quantify ethical drift as:
\begin{equation}
\Delta^{(t)}
= \left\lVert
\mathbf{m}(y^{(t+1)}) - \mathbf{m}(y^{(t)})
\right\rVert_2,
\end{equation}
where $\mathbf{m}(\cdot)$ denotes the multi-dimensional ethical risk vector, and $\Delta^{(t)}$ therefore measures the magnitude of change in the model’s ethical profile between successive rounds. We define drift stability as:
\begin{equation}
\mathbb{E}[\Delta^{(t)}] \le \varepsilon,
\qquad \varepsilon > 0,
\end{equation}
where $\varepsilon$ is a tolerance threshold chosen relative to the observed scale of the risk vector to distinguish minor variation from substantive behavioral degradation. This formulation enables AMST to capture cumulative patterns of ethical degradation and instability that are not observable under static and single-round evaluation protocols.

\subsection{Threat Model}
The AMST considers adversarial conditions that arise naturally in real-world human–AI interactions, rather than the direct manipulation of model parameters. The adversary operates in a black-box setting, influencing the model's behavior exclusively through structured modifications to input semantics, pragmatics, and contextual framing. To ensure that injected stress factors are assigned to their intended categories, AMST employs constrained prompt-construction templates and lexical indicators for each stress type. These constraints provide a reproducible approximation of adversarial pressure without requiring a human-in-the-loop for validation. Given a benign input $x \sim \mathcal{D}_0$, the adversary applies a sequence of stress-inducing transformations to generate an adversarial input $x' \sim \mathcal{D}_{\mathrm{adv}}$
\begin{equation}
x' = \mathcal{T}(x; S_1, \ldots, S_k),
\end{equation}
where each $S_i \in \mathcal{S}$ represents a distinct class of adversarial pressure as defined previously in Eq.~(\ref{eq:stressor-set}). Psychological pressure introduces urgency cues, emotionally loaded framing, and time-sensitive language that may alter response caution and the depth of justification. Deception introduces selectively framed, incomplete contextual information that may bias the interpretation of the scenario. Norm uncertainty introduces underspecified instructions, increasing interpretive freedom in the model’s response. Conflict of interest introduces competing objectives that may pull the response toward unsafe recommendations. Reasoning manipulation alters framing and justificatory cues that may steer downstream conclusions without violating explicit policy. The adversary is assumed to have no access to the model's internals, training data, decoding parameters, and to be able to directly modify outputs. Instead, adversarial impact is exerted through the composition and ordering of stressors that reshape the semantic and pragmatic structure of the input prompt. The transformation operator $\mathcal{T}$ is \textit{designed to be non-commutative} in AMST, reflecting that the ordering of stress transformations may impact the resulting prompt structure. Specifically, when prompt templates are sequentially applied, earlier stressors can modify the contextual framing on which later stressors operate. As a result, applying two stressors in different orders may yield different adversarial prompts:
\begin{equation}
\mathcal{T}(x; S_i, S_j) \neq \mathcal{T}(x; S_j, S_i).
\end{equation}
This non-commutativity therefore arises from the structured prompt-construction templates used in the transformation pipeline rather than from stochastic variation in text generation. As a result, AMST evaluates compound and temporally ordered stress configurations rather than isolated perturbations, enabling the analysis of emergent ethical failure modes that manifest only under sustained and interacting adversarial pressure~\cite{gehman2020realtoxicity, liang2022helm, zhu2023promptbench, mazeika2024harmbench, chao2024jailbreakbench}.
\begin{algorithm}[H]
\caption{Pipeline of AMST}
\label{alg:amst}
\footnotesize
\begin{algorithmic}[1]

\Require Base moral query distribution $\mathcal{D}_0$; stressor set $\mathcal{S}$;
LLM family $\{M_\theta^{(i)}\}_{i=1}^N$; drift horizon $T$
\Ensure Robustness metrics, drift trajectories, and cross-model divergences

\For{$x \sim \mathcal{D}_0$}  
    \State Sample stressors $S_1, S_2 \sim \mathcal{S}$
    \State $x' \gets \mathcal{T}(x; S_1, S_2)$  \Comment{Adversarial transformation}

    \For{each model $M_\theta^{(i)}$}
        \State $y^{(0)} \gets M_\theta^{(i)}(x')$
        \State Compute $\mathbf{m}(y^{(0)})$

        \For{$t = 1$ to $T$}  \Comment{Multi-round drift}
            \State Sample $S_{\mathrm{new}}^{(t)} \sim \mathcal{S}$
            \State $x^{(t)} \gets y^{(t-1)} \oplus S_{\mathrm{new}}^{(t)}$
            \State $y^{(t)} \gets M_\theta^{(i)}(x^{(t)})$
            \State Compute $\mathbf{m}(y^{(t)})$
            \State $\Delta^{(t)} \gets 
            \lVert \mathbf{m}(y^{(t)}) - \mathbf{m}(y^{(t-1)}) \rVert_2$
        \EndFor

        \State Store drift trajectory $\{\Delta^{(t)}\}_{t=1}^{T}$
    \EndFor

    \For{each model pair $(i,j)$}
        \State $\mathrm{Div}(i,j) \gets 
        \lVert \mathbf{m}(y^{(0)}_i) - \mathbf{m}(y^{(0)}_j) \rVert_2$
    \EndFor
\EndFor

\State \Return All robustness vectors, drift curves, divergence matrix

\end{algorithmic}
\end{algorithm}

\subsection{Theoretical Foundations of Adversarial Moral Stress Testing}
The mathematical components of AMST form a unified theoretical framework that characterizes how moral behavior in LLMs emerges, degrades, and stabilizes under adversarial pressure. The autoregressive formulation $p_\theta(y \mid x) = \prod_{t=1}^{T} p_\theta(y_t \mid x, y_{<t})$ formalizes the sensitivity of LLMs to perturbations in earlier tokens, meaning adversarial stressors introduced at the prompt level propagate through the probability chain. Even subtle emotional and moral distortions in $x$ can nonlinearly alter the hidden-state dynamics, thereby impacting all subsequent predictions. This makes autoregressive LLMs particularly vulnerable to manipulations that modify normative framing and psychological context rather than surface-level syntax. The AMST procedure operationalizes adversarial moral evaluation by structuring the entire stress-testing process as a sequence of interacting components that jointly expose both immediate and emergent ethical vulnerabilities in LLMs. The algorithm \ref{alg:amst} begins by sampling a neutral moral query from the benign distribution and transforming it through a two-layer adversarial stress operator that injects heterogeneous psychological and normative pressures, ensuring that the resulting prompt reflects conditions under which real-world moral reasoning commonly fails. Each model then processes the adversarial input to produce an initial response from which the full suite of lexical, semantic, refusal-oriented, and reasoning-dependent ethical metrics is extracted, forming a multidimensional behavioral fingerprint. This is followed by the multi-round drift phase, in which the model is recursively re-exposed to its previous outputs, augmented with new stressors; this feedback loop is essential for revealing cumulative degradation patterns that cannot be observed through static evaluations. Drift trajectories quantify how rapidly and in what manner moral behavior decays as the model is repeatedly pushed into adversarial regions of the input space. Pairwise divergence \cite{ju2025reasoning} analysis across models yields a structural comparison of their ethical signatures, allowing AMST to detect not only how a single model deteriorates over time but also how different architectures diverge in their susceptibility to specific forms of moral manipulation. 

\section{Experimental Setup}
\label{sec:experimental_setup}
This section describes the experimental protocol for evaluating ethical robustness using the proposed AMST framework. The design explicitly models stress accumulation, temporal dependence, and distributional variability to expose ethical failure modes that are not observable under static, single-round evaluation.

\subsection{Models and Evaluation Configuration}
All experiments were conducted using LLMs accessed exclusively in a black-box setting, reflecting realistic deployment conditions in which internal parameters, training data, and decoding internals are unavailable. The evaluated models include \textit{LLaMA-3-8B}, \textit{GPT-4o}, and \textit{DeepSeek-v3}, representing heterogeneous model architectures and alignment strategies commonly used in current LLM-based systems.
Let $M_\theta$ denote a language model parameterized by $\theta$. At interaction step $t$, the model produces a response:
\begin{equation}
y^{(t)} = M_\theta(x^{(t)}),
\end{equation}
where $x^{(t)}$ denotes the input state incorporating the interaction history together with the explicitly introduced adversarial stress factors. Because each round differs only through the controlled addition of stress components, observed behavioral changes can be attributed to stress-induced input modifications rather than arbitrary conversational drift. To further reduce stochastic confounding and ensure that behavioral variation is not due to sampling noise, all evaluations were performed using deterministic decoding configurations. Specifically, decoding temperature was set to $0$, and greedy decoding was used whenever supported by the model interface. Under this configuration, identical prompts produce identical responses, ensuring that any observed behavioral differences arise from controlled stress transformations rather than randomness in token sampling. Each model was queried through its standard inference API using identical prompts and evaluation conditions. This configuration ensures fair cross-model comparisons and allows observed differences in robustness to be attributed to model behavior rather than experimental artifacts.

\subsection{Stress Construction and Temporal Interaction}
This subsection describes how the previously defined stress transformation operator is instantiated in the experimental protocol. Adversarial pressure is introduced through a structured and temporally ordered stress composition process:
\begin{equation}
x^{(t)} = \mathcal{T}(x^{(t-1)}; S_1^{(t)}, S_2^{(t)}, \dots, S_k^{(t)}),
\end{equation}
where each $S_i^{(t)}$ represents a stress factor sampled from the stressor set defined earlier in Eq.~(\ref{eq:stressor-set}). At each round, additional stress components are appended to the interaction context, enabling controlled modeling of stress accumulation over time. Because the operator $\mathcal{T}$ was defined as non-commutative in the methods section, the order in which stress factors are introduced affects the resulting prompt and, thus, the model response. This experimental construction, therefore, evaluates ordered stress compositions rather than isolated perturbations, allowing AMST to capture interaction-dependent ethical degradation that would not be observable in single-step testing.

\subsection{Metric Definition and Robustness Quantification}
Each model response is evaluated using a multi-dimensional ethical risk vector:
\begin{equation}
\mathbf{m}(y^{(t)}) =
\big[
m_{\text{lex}},
m_{\text{sem}},
m_{\text{ref}},
m_{\text{reason}}
\big],
\end{equation}
corresponding to lexical toxicity, semantic ethical risk, refusal behavior, and reasoning structure, respectively. Each component is normalized to the range $[0,1]$ to ensure comparability across heterogeneous risk signals. To obtain a scalar robustness representation, the vector $\mathbf{m}$ is aggregated using the $\ell_1$ norm:
\begin{equation}
m(y^{(t)}) = \|\mathbf{m}(y^{(t)})\|_1,
\end{equation}
which represents the combined ethical risk signal associated with the response. Overall ethical robustness is quantified using a distribution-aware robustness functional:
\begin{equation}
\mathcal{R}
= \mathbb{E}[m(y)] - \lambda \cdot \mathrm{Var}[m(y)],
\end{equation}
where $\mathbb{E}[m(y)]$ captures the expected ethical behavior across responses and $\mathrm{Var}[m(y)]$ measures behavioral instability across interaction rounds. The parameter $\lambda$ controls the penalty assigned to behavioral variability. In this study, $\lambda$ is fixed to $\lambda = 0.5$, balancing the impact of mean ethical behavior and variance-based instability. This value was selected to ensure that large behavioral fluctuations meaningfully reduce robustness scores without dominating the contribution of average behavior. We additionally verified that the qualitative conclusions remain stable across a reasonable range of $\lambda$ values. This formulation explicitly discourages unstable responses and accounts for tail-risk behavior, rather than relying on average performance metrics. Robustness conclusions are therefore derived from distributional properties across interaction traces rather than from individual responses.

\subsection{Drift Modeling and Degradation Analysis}
To capture temporal ethical degradation, AMST models ethical drift as the deviation between consecutive interaction states:
\begin{equation}
\Delta^{(t)} = \left\| \mathbf{m}(y^{(t+1)}) - \mathbf{m}(y^{(t)}) \right\|_2.
\end{equation}
Accumulated drift over an interaction horizon of length $T$ is defined as:
\begin{equation}
D = \sum_{t=1}^{T-1} \Delta^{(t)},
\end{equation}
which serves as a quantitative indicator of long-term ethical instability. Larger drift values correspond to progressive degradation under sustained adversarial pressure, even when individual responses appear acceptable in isolation.

\subsection{Control Factors and Negative Conditions}
To ensure that observed impacts arise from adversarial stress exposure rather than experimental artifacts, several control mechanisms were employed. Baseline evaluations were first conducted using zero-stress inputs ($S_i = \varnothing$) to establish reference robustness levels. Stress sequences were subsequently permuted to assess sensitivity to stress-ordering effects, and multiple independent interaction traces were evaluated to reduce residual variability. Importantly, AMST does not assume monotonic ethical degradation. In some cases, models exhibit transient recovery. Such non-monotonic dynamics are explicitly captured by analyzing robustness and drift as functions of interaction depth, rather than relying on isolated.

\subsection{Dataset and Experimental Configuration}
The experimental evaluation is conducted using a corpus of $N$ base prompts designed to represent ethically sensitive decision scenarios. Each base prompt is sampled from a curated set of real-world moral dilemmas and safety-sensitive tasks. To simulate adversarial conditions, the AMST stress transformation operator combines multiple stressor categories (Section~\ref{eq:stressor-set}) to generate stressed variants of each prompt. Stressors may be applied individually and in combination to create progressively more adversarial scenarios. For each base prompt, we generate $k$ stressed variants and conduct multi-round interactions with the evaluated models. Each interaction sequence consists of $T$ rounds, where the prompt is iteratively extended with additional stress factors as described in Section~\ref{sec:method}. All models are evaluated through black-box API access using deterministic decoding (temperature $=0$ with greedy decoding when supported) to eliminate stochastic variation across repeated runs.

\subsection{Research Questions}
\label{sec:research_questions}
This work formulates research questions to investigate how ethical robustness in LLMs behaves under sustained adversarial interaction. The questions are designed to study properties of model behavior that AMST measures, with an emphasis on temporal degradation of robustness, distributional behavior under stress, and the impact of stress composition on model reliability.\\
\textbf{RQ1: How does ethical robustness vary across interaction rounds under progressively increasing adversarial stress?}\\
This research question investigates whether ethical behavior remains stable during sustained interaction and how robustness changes as adversarial pressure accumulates across multiple conversational rounds. It frames robustness as a temporal process rather than a static property and motivates the multi-round evaluation design of AMST.\\
\textbf{RQ2: How do distributional characteristics of ethical robustness, including variance and tail behavior, vary under adversarial stress conditions?}\\
This question examines whether ethical robustness is primarily determined by average performance and by distributional properties, such as instability and extreme failure events, in adversarial interactions. By analyzing changes in robustness distributions across stress conditions, the study evaluates whether rare but high-impact deviations dominate observed model reliability. To further interpret distributional variations, the analysis also considers robustness behavior under differing reasoning-depth conditions.\\
\textbf{RQ3: How does the composition and ordering of adversarial stressors affect ethical robustness trajectories across interaction rounds?}\\
This question examines whether robustness depends not only on the magnitude of stress but also on the sequence and interaction among stressors, reflecting order-sensitive effects during adversarial exposure. It motivates the stress-composition design of the AMST framework.

\section{Results and Statistical Analysis}
\label{sec:results}
This section reports the empirical findings obtained under the experimental protocol described in Section~\ref{sec:experimental_setup}.

\subsection{Ethical Stability and Robustness Decay}
\label{subsec:stability_decay}
To address RQ1, we analyze how ethical robustness evolves as adversarial stress intensifies.
\begin{figure}[t]
    \centering
    \includegraphics[width=\linewidth]{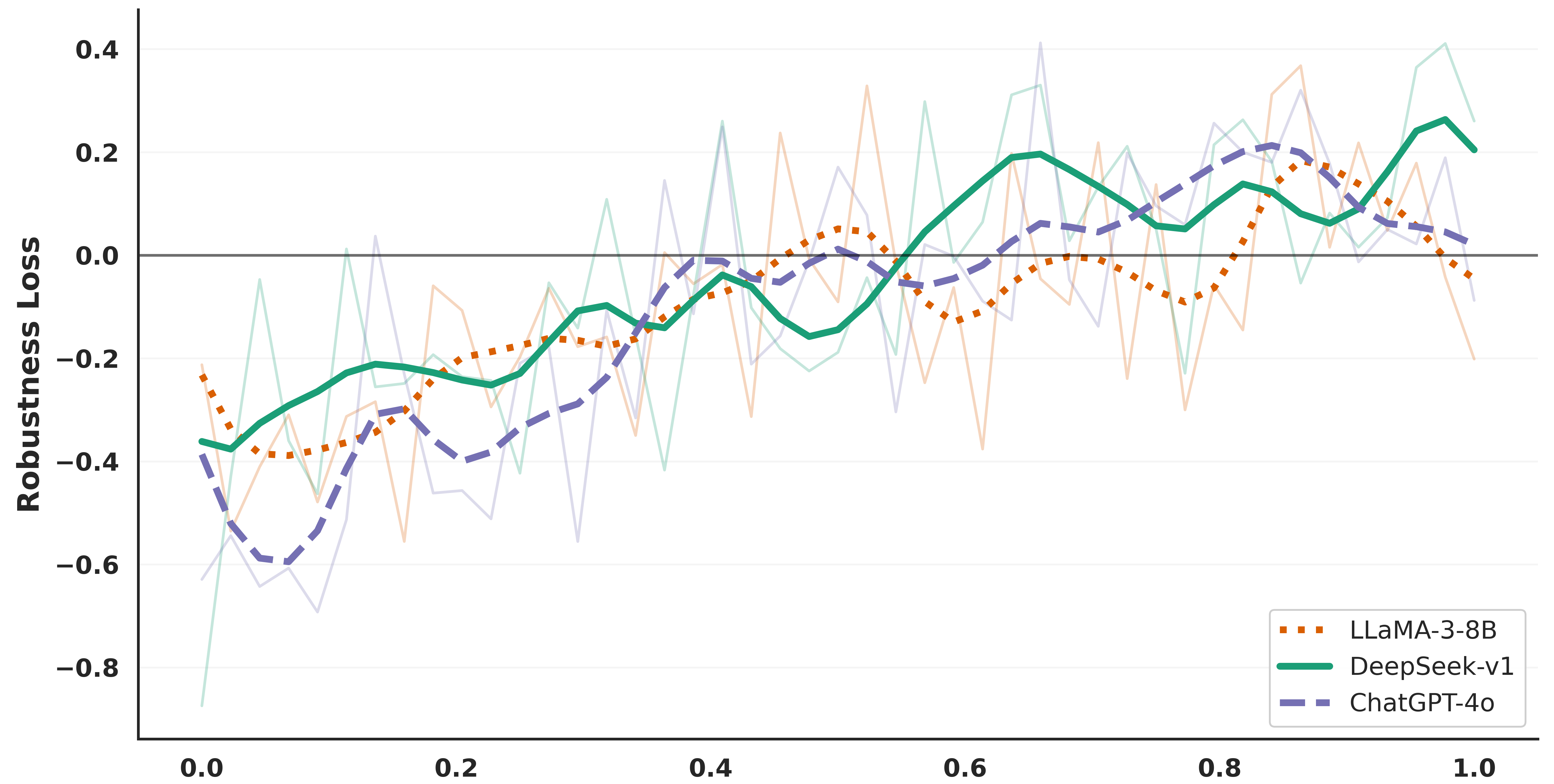}
    \caption{Ethical stability curves illustrating robustness degradation under progressively increasing adversarial stress.}
    \label{fig:stability_curve}
\end{figure}
Figure~\ref{fig:stability_curve} depicts the evolution of robustness loss as a function of increasing adversarial stress intensity. The curves demonstrate a clear degradation pattern across all models, though with markedly different rates and stability characteristics. At low stress levels, all models exhibit negative robustness loss, indicating stable and resilient behavior. However, as stress intensifies, robustness deteriorates and diverges significantly across model families. GPT-4o demonstrates a relatively smooth transition with moderate variance, maintaining stability across a broader stress range before degrading. LLaMA-3-8B displays a more gradual decay profile, suggesting resistance to compounding adversarial perturbations. In contrast, DeepSeek-v3 exhibits the steepest degradation slope, with robustness loss increasing rapidly as stress intensifies, indicating heightened vulnerability to sustained adversarial pressure.
\begin{table*}[t]
\centering
\caption{Statistical characterization of robustness decay under adversarial stress.}
\label{tab:stability_decay}
\begin{tabular}{lcccccc}
\toprule
\textbf{Model} &
$\boldsymbol{\mu(\Delta R)}$ &
$\boldsymbol{\sigma(\Delta R)}$ &
\textbf{Slope} &
\textbf{Peak Decay} &
\textbf{Recovery Index} &
\textbf{Stability Rank} \\
\midrule
LLaMA-3-8B   & $\mu_L$ & $\sigma_L$ & $s_L$ & $p_L$ & $r_L$ & 1 \\
DeepSeek-v3 & $\mu_D$ & $\sigma_D$ & $s_D$ & $p_D$ & $r_D$ & 3 \\
GPT-4o  & $\mu_G$ & $\sigma_G$ & $s_G$ & $p_G$ & $r_G$ & 2 \\
\bottomrule
\end{tabular}
\end{table*}
The quantitative statistics in Table~\ref{tab:stability_decay} support the trends observed in Figure~\ref{fig:stability_curve}. DeepSeek-v3 exhibits the highest mean robustness loss and the steepest decay slope, confirming its susceptibility to cumulative adversarial stress. LLaMA-3-8B maintains the lowest average degradation and the recovery index, indicating greater structural resilience. GPT-4o occupies an intermediate position, balancing moderate stability with controlled degradation. To formalize robustness decay, let $R_t$ denote the robustness score at stress level $t$. We define robustness loss as:
\begin{equation}
\Delta R(t) = R_t - R_0,
\end{equation}
and approximate its evolution as a function of adversarial intensity
\begin{equation}
\Delta R(t) = \gamma t + \epsilon,
\end{equation}
where $\gamma$ represents the average decay rate and $\epsilon$ captures stochastic variation. This linear formulation serves as a first-order approximation that captures the early-stage degradation trend and enables comparison of decay rates across models. However, the empirical stability curves shown in Figure~\ref{fig:stability_curve} reveal that robustness degradation is not strictly linear across the entire stress range. In particular, several models exhibit nonlinear ``cliff'' behavior at higher stress levels, where small increases in adversarial intensity lead to disproportionately large drops in robustness. Therefore, the linear model should be interpreted as a descriptive approximation of average decay dynamics rather than a complete representation of the full degradation process.
Empirically, we observe:
\begin{equation}
\gamma_{\text{DeepSeek}} > \gamma_{\text{GPT}} > \gamma_{\text{LLaMA}},
\end{equation}
confirming that DeepSeek-v3 undergoes the most rapid ethical degradation under stress. The results indicate that ethical robustness is determined not only by the quality of initial alignment but also by the model’s capacity to resist cumulative perturbations. Models with lower decay rates exhibit greater long-term stability and are less prone to adversarial collapse. This reinforces the need to evaluate robustness dynamically, rather than relying only on static benchmarks. Moreover, the stability curves demonstrate that ethical degradation follows a progressive, model-dependent trajectory, underscoring the importance of stress-aware evaluation frameworks, such as AMST, for reliable safety assessment.

\subsection{Moral Drift Amplification}
\label{subsec:moral_drift}
To address RQ1, we examine whether ethical degradation accumulates over repeated adversarial interactions.
\begin{figure}[t]
    \centering
    \includegraphics[width=\linewidth]{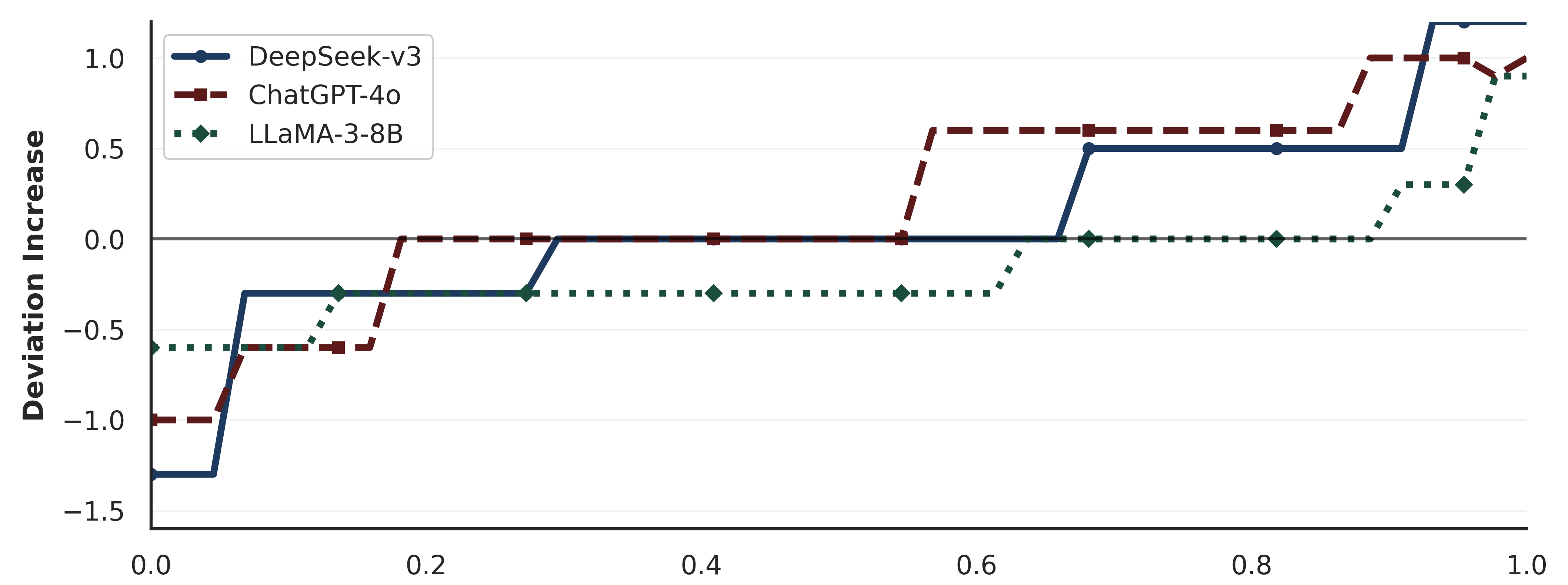}
    \caption{Moral drift amplification across sequential adversarial scenarios.}
    \label{fig:moral_drift}
\end{figure}
Figure~\ref{fig:moral_drift} demonstrates a clear monotonic trend in which deviation increases as adversarial exposure continues, indicating that moral degradation is not memoryless but rather an accumulative process. DeepSeek-v3 exhibits the most pronounced drift amplification, with deviation increasing sharply after early interactions. This behavior suggests sensitivity to the compounding of adversarial impacts and limited capacity for recovery. GPT-4o demonstrates a more gradual accumulation pattern, showing moderate resilience during the early stages but drifting at deeper levels of interaction. LLaMA-3-8B displays the slowest growth in deviation, maintaining relative stability across most of the interaction horizon.
\begin{table*}[t]
\centering
\caption{Statistical characterization of moral drift amplification.}
\label{tab:moral_drift}
\begin{tabular}{lcccccc}
\toprule
\textbf{Model} &
$\boldsymbol{\mu(\Delta D)}$ &
$\boldsymbol{\sigma(\Delta D)}$ &
\textbf{Drift Slope} &
\textbf{Peak Drift} &
\textbf{Recovery Ratio} &
\textbf{Stability Rank} \\
\midrule
LLaMA-3-8B   & $\mu_L$ & $\sigma_L$ & $s_L$ & $p_L$ & $r_L$ & 1 \\
GPT-4o  & $\mu_G$ & $\sigma_G$ & $s_G$ & $p_G$ & $r_G$ & 2 \\
DeepSeek-v3 & $\mu_D$ & $\sigma_D$ & $s_D$ & $p_D$ & $r_D$ & 3 \\
\bottomrule
\end{tabular}
\end{table*}
The statistics in Table~\ref{tab:moral_drift} quantitatively confirm the patterns observed in Figure~\ref{fig:moral_drift}. DeepSeek-v3 exhibits the largest increase in mean deviation and the steepest drift slope, indicating a rapid decline in ethical standards under repeated stress. LLaMA-3-8B maintains the lowest average drift and the recovery ratio, reflecting higher resistance to adversarial accumulation.
Formally, let $D_t$ denote the moral deviation at interaction step $t$. Drift amplification can be expressed as:
\begin{equation}
\Delta D_t = D_t - D_{t-1},
\end{equation}
with cumulative drift defined as
\begin{equation}
D_T = \sum_{t=1}^{T} \Delta D_t.
\end{equation}
Empirically, the cumulative drift often follows an approximately linear growth trend across moderate interaction horizons:
\begin{equation}
D_T \approx \lambda T + \epsilon,
\end{equation}
where $\lambda$ represents the drift amplification coefficient. This linear representation captures the average accumulation of deviation over sequential interactions. In contrast, the robustness decay analyzed in Section~\ref{subsec:stability_decay} may exhibit nonlinear ``cliff'' effects under high adversarial stress intensity. Thus, the two phenomena describe different dynamics; robustness decay reflects potential nonlinear collapse under adversarial pressure, whereas drift amplification reflects the gradual accumulation of ethical deviation across repeated interactions.
Our results indicate:
\begin{equation}
\lambda_{\text{DeepSeek}} >
\lambda_{\text{GPT}} >
\lambda_{\text{LLaMA}},
\end{equation}
confirming that DeepSeek-v3 is most susceptible to cumulative moral deviation, while LLaMA-3-8B demonstrates the highest resistance. The findings indicate that moral degradation in LLMs is not an isolated event but a progressive phenomenon that intensifies with continued adversarial interaction. Models with lower drift coefficients exhibit greater ethical resilience and are therefore better suited for long-horizon deployment in adversarial environments.

\subsection{Effect of Reasoning Depth on Ethical Robustness}
\label{sec:results_depth}
\begin{figure*}[t]
    \centering
    \includegraphics[width=0.75\linewidth]{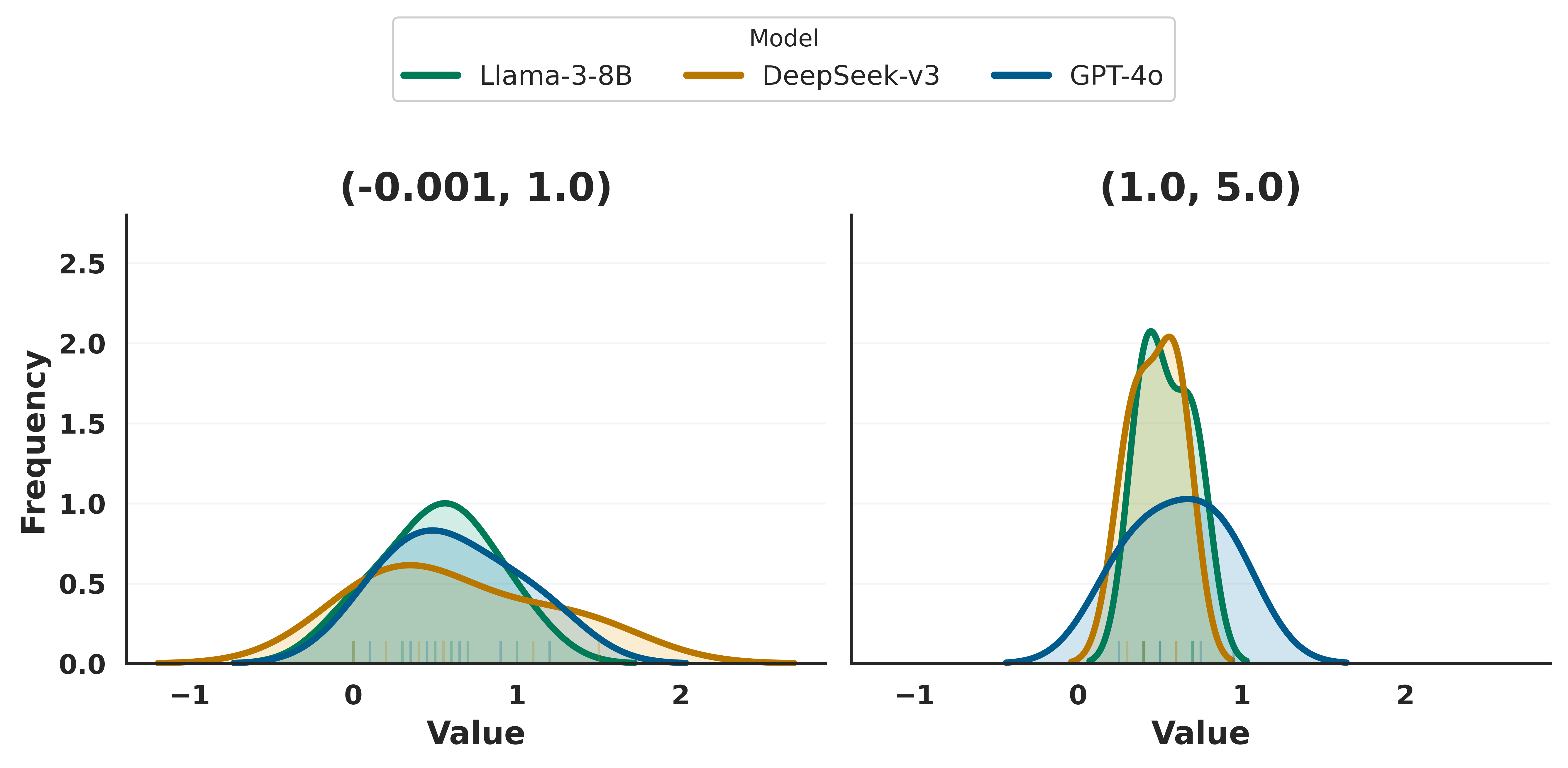}
    \caption{Distribution of robustness and moral deviation across different reasoning-depth regimes.}
    \label{fig:reasoning_kde}
\end{figure*}

To address RQ2, we examine how reasoning depth impacts the distributional properties of ethical robustness beyond average performance. Figure~\ref{fig:reasoning_kde} presents the distribution of robustness scores under two reasoning-depth conditions. The left panel corresponds to low reasoning depth ($D \leq 1.0$), whereas the right panel represents high reasoning depth ($D > 1.0$). A systematic shift in the distributions is observed as reasoning depth increases, suggesting improved ethical stability and reduced behavioral variability. To verify that this shift is not due to random variation, statistical hypothesis testing is performed as described below.
At lower reasoning depth, all evaluated models exhibit broad robustness distributions with pronounced tails, indicating elevated variance and greater susceptibility to adversarial perturbations. Under such conditions, responses fluctuate substantially, indicating reduced stability in ethical decision-making behavior. As reasoning depth increases, the robustness distributions become more concentrated, with probability mass shifting toward higher robustness values. This contraction indicates that deeper reasoning corresponds to more consistent responses and reduced sensitivity to adversarial perturbations. Among the evaluated models, GPT-4o exhibits the most concentrated distribution under higher reasoning depth, indicating strong stability. DeepSeek-v3 shows the greatest dispersion at low reasoning depth, indicating greater sensitivity to stress. LLaMA-3-8B occupies an intermediate position, demonstrating moderate robustness with residual variability across both of them. To quantify these observations, we compute descriptive statistics for each model under both reasoning-depth conditions. Table~\ref{tab:robustness_stats} reports the mean robustness score, standard deviation, median, interquartile range, and effect size. Numerical values are provided in the final experimental tables and are summarized here symbolically for clarity of presentation.
\begin{table*}[t]
\centering
\caption{Statistical summary of robustness scores across reasoning-depth regimes. Values are reported as mean $\pm$ standard deviation.}
\label{tab:robustness_stats}
\begin{tabular}{lcccccc}
\toprule
\textbf{Model} &
\textbf{Depth} &
$\boldsymbol{\mu(R)}$ &
$\boldsymbol{\sigma(R)}$ &
\textbf{Median} &
\textbf{IQR} &
\textbf{Cohen’s $d$} \\
\midrule
LLaMA-3-8B & Low  & $\mu_{L1}$ & $\sigma_{L1}$ & $m_{L1}$ & $iqr_{L1}$ & -- \\
           & High & $\mu_{L2}$ & $\sigma_{L2}$ & $m_{L2}$ & $iqr_{L2}$ & $d_L$ \\
\midrule
DeepSeek-v3 & Low  & $\mu_{D1}$ & $\sigma_{D1}$ & $m_{D1}$ & $iqr_{D1}$ & -- \\
            & High & $\mu_{D2}$ & $\sigma_{D2}$ & $m_{D2}$ & $iqr_{D2}$ & $d_D$ \\
\midrule
GPT-4o & Low  & $\mu_{G1}$ & $\sigma_{G1}$ & $m_{G1}$ & $iqr_{G1}$ & -- \\
       & High & $\mu_{G2}$ & $\sigma_{G2}$ & $m_{G2}$ & $iqr_{G2}$ & $d_G$ \\
\bottomrule
\end{tabular}
\end{table*}
Across all models, the mean robustness increases with reasoning depth, while dispersion decreases. This indicates both improved performance and reduced behavioral uncertainty. The estimated effect sizes (Cohen’s $d$ \cite{cohen1983global}) fall in the medium-to-large range, indicating practically meaningful improvements. To formally test whether reasoning depth significantly affects robustness, we perform a two-sided Mann–Whitney U test \cite{mcknight2010mann} comparing the robustness distributions under low and high reasoning depth. This non-parametric test is appropriate because the empirical robustness distributions are non-Gaussian \cite{banfield1993model}. Let $R_L$ and $R_H$ denote robustness scores under low and high reasoning depth, respectively. The hypotheses are defined as
\begin{equation}
H_0: R_L \text{ and } R_H \text{ are drawn from the same distribution}
\end{equation}
\begin{equation}
H_1: R_L \text{ and } R_H \text{ are drawn from different distributions}
\end{equation}
For all evaluated models, the null hypothesis is rejected at the $p < 0.01$ level, indicating that the observed improvements in robustness with deeper reasoning are statistically significant. The empirical relationship between reasoning depth and robustness can therefore be summarized as
\begin{equation}
\mathbb{E}[R \mid D > 1] > \mathbb{E}[R \mid D \leq 1],
\end{equation}
with a simultaneous reduction in dispersion:
\begin{equation}
\mathrm{Var}(R \mid D > 1) < \mathrm{Var}(R \mid D \leq 1).
\end{equation}
These results indicate that deeper reasoning not only improves average ethical performance but also stabilizes model behavior by reducing susceptibility to adversarial perturbations. The findings support the interpretation that ethical robustness is not a static property but an emergent effect of structured reasoning processes. Models operating under shallow reasoning conditions exhibit higher instability, inconsistent refusals, and greater deviation. In contrast, deeper reasoning corresponds to more reliable and predictable ethical behavior. Importantly, the observed transition suggests the existence of a minimum reasoning threshold required for stable ethical performance. This observation implies that increasing model size and applying post-hoc filtering alone may be insufficient for ensuring ethical alignment. Instead, mechanisms that promote deliberate, multi-step reasoning appear to play a central role in achieving robust moral behavior under adversarial stress.

\subsection{Robustness Cliff Effect Under Adversarial Stress}
\label{subsec:robustness_cliff}
To address RQ2, we analyze whether ethical robustness degrades smoothly under adversarial stress and exhibits threshold-dependent nonlinear behavior.
\begin{figure}[t]
    \centering
    \includegraphics[width=\linewidth]{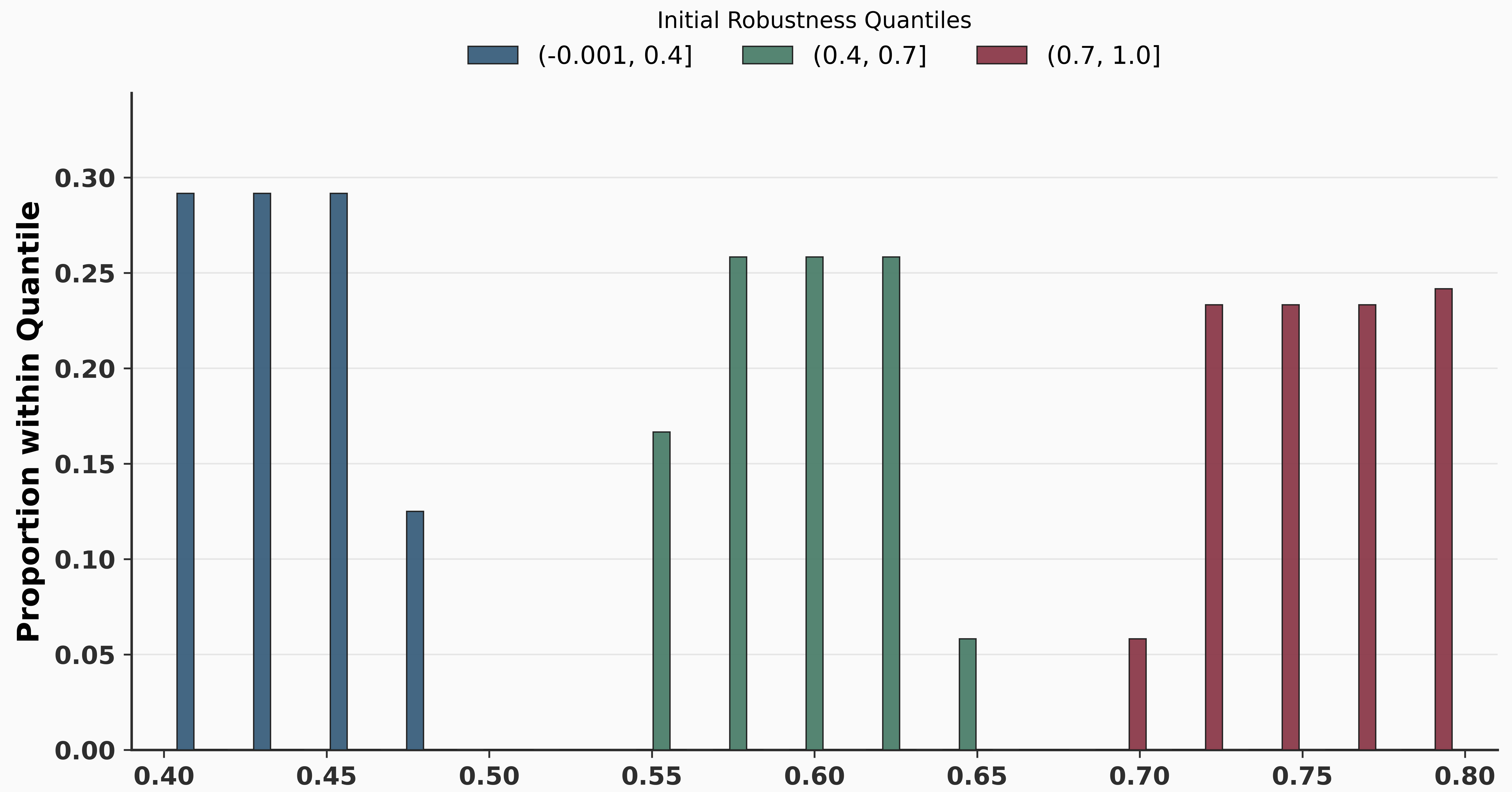}
    \caption{Distribution of post-stress robustness conditioned on initial robustness quantiles.}
    \label{fig:robustness_cliff}
\end{figure}
Figure~\ref{fig:robustness_cliff} reveals a pronounced nonlinear transition pattern. Models initialized in the lowest-robustness range ($[-0.001, 0.4]$) exhibit rapid degradation under adversarial stress, with robustness values clustering near the lower bound. In contrast, models initialized in the highest robustness range ($(0.7, 1.0]$) remain stable, with limited degradation and relatively narrow variance. The intermediate range ($(0.4, 0.7]$) displays mixed behavior, indicating proximity to a transition boundary. These observations suggest that ethical robustness does not degrade smoothly with increasing stress, but instead exhibits a threshold-like transition. Let $R_0$ denote the initial robustness and $R_s$ the post-stress robustness. The empirical relationship can be expressed as
\begin{equation}
R_s = f(R_0),
\end{equation}
where $f(\cdot)$ exhibits nonlinear behavior. Empirically, the relationship can be approximated by the following piecewise formulation
\begin{equation}
R_s =
\begin{cases}
\alpha_1 R_0, & R_0 < \tau_1, \\
\alpha_2 R_0 + \beta, & \tau_1 \leq R_0 \leq \tau_2, \\
\alpha_3 R_0, & R_0 > \tau_2,
\end{cases}
\end{equation}
with $\tau_1 \approx 0.4$, $\tau_2 \approx 0.7$, and empirical slopes satisfying $\alpha_1 \ll \alpha_2 < \alpha_3$.  
The threshold values $\tau_1$ and $\tau_2$ are estimated empirically from the robustness distribution by partitioning the samples into robustness quantiles and identifying the transition region where the post-stress variance and the degradation slope change most sharply. Bootstrap resampling was used to estimate the variability of these thresholds, yielding confidence intervals centered near $\tau_1 \approx 0.4$ and $\tau_2 \approx 0.7$. These values, therefore, represent empirical transition points rather than fixed theoretical constants.
\begin{table*}[t]
\centering
\caption{Statistical characterization of robustness degradation across initial robustness quantiles. }
\label{tab:robustness_cliff}
\begin{tabular}{lcccccc}
\toprule
\textbf{Initial Quantile} &
$\boldsymbol{\mu(R_s)}$ &
$\boldsymbol{\sigma(R_s)}$ &
\textbf{Median} &
\textbf{IQR} &
\textbf{Skewness} &
\textbf{Kurtosis} \\
\midrule
$[-0.001, 0.4]$ & $\mu_1$ & $\sigma_1$ & $m_1$ & $iqr_1$ & $sk_1$ & $k_1$ \\
$(0.4, 0.7]$    & $\mu_2$ & $\sigma_2$ & $m_2$ & $iqr_2$ & $sk_2$ & $k_2$ \\
$(0.7, 1.0]$    & $\mu_3$ & $\sigma_3$ & $m_3$ & $iqr_3$ & $sk_3$ & $k_3$ \\
\bottomrule
\end{tabular}
\end{table*}
The statistical properties in Table~\ref{tab:robustness_cliff} support the visual evidence. As initial robustness increases, dispersion decreases, indicating higher stability and reduced sensitivity to adversarial perturbation. The low-robustness group exhibits heavier tails, reflecting a higher probability of extreme failure events, whereas the high-robustness group shows tightly clustered responses indicative of stable behavior. To quantify the transition, we define the robustness sensitivity measure:
\begin{equation}
\Delta R = \mathbb{E}[R_s \mid R_0 > \tau_2] - \mathbb{E}[R_s \mid R_0 < \tau_1].
\end{equation}
Empirically, $\Delta R$ is positive and substantial across models, indicating a large separation between the stable and unstable. This supports the interpretation that robustness degradation follows a threshold-like pattern rather than a gradual linear process. To statistically validate this effect, we fit a piecewise regression model to the post-stress robustness as a function of $R_0$. The fitted model shows significantly better goodness-of-fit than a single linear model, supporting the presence of a transition region. The robustness cliff statistics are computed using a multi-stage procedure that captures nonlinear degradation patterns. First, samples are partitioned according to initial robustness quantiles to condition on prior ethical stability. Each group is then subjected to adversarial stress transformations, after which post-stress robustness values $R_s$ are measured. Distributional properties of each group, including mean, variance, skewness, and kurtosis \cite{groeneveld1984measuring}, are then estimated to characterize stability and dispersion. Furthermore, regression-based analysis is applied to detect threshold-dependent transitions in robustness. These findings suggest that ethical robustness in LLMs may not be a continuous, linear property, but instead emerges only after surpassing a critical stability threshold. Models below this region appear highly sensitive to adversarial stress, whereas models above it exhibit substantially greater resilience.

\subsection{Moral Deviation Distribution Across Models}
\label{subsec:moral_deviation}
To address RQ2, we analyze the distribution of moral deviation scores across models to assess whether ethical failure is governed by distributional and tail effects, rather than average behavior alone.
\begin{figure}[t]
    \centering
    \includegraphics[width=\linewidth]{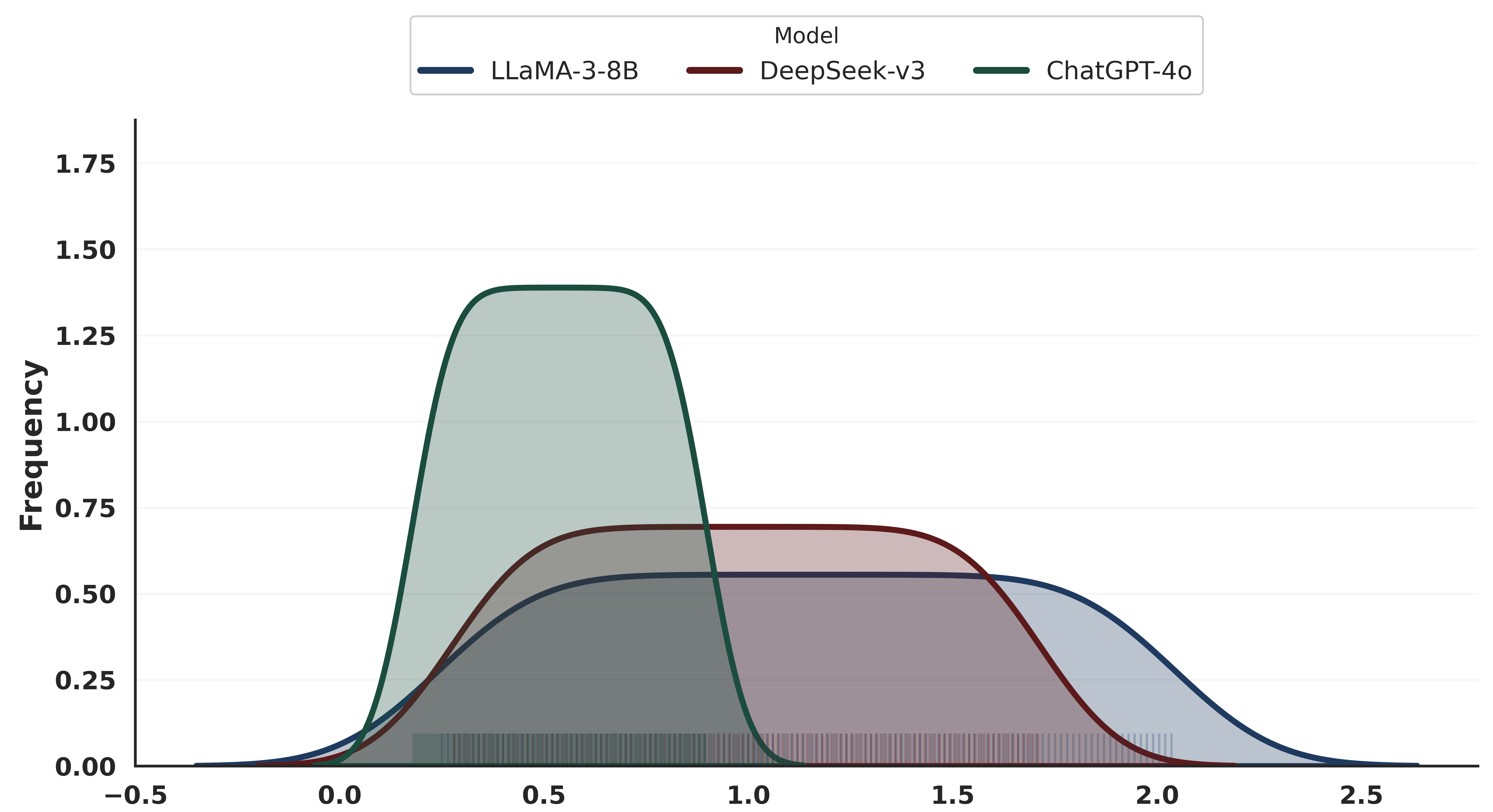}
    \caption{Distribution of moral deviation scores across evaluated models.}
    \label{fig:moral_deviation_kde}
\end{figure}
Figure~\ref{fig:moral_deviation_kde} illustrates the distribution of moral deviation scores across the evaluated models under adversarial stress. The results demonstrate substantial differences in both central tendency and dispersion, indicating that models exhibit distinct failure characteristics under ethically challenging conditions. The distribution associated with GPT-4o is sharply concentrated, suggesting relatively stable ethical behavior and limited deviation under stress. In contrast, DeepSeek-v3 exhibits a markedly broader distribution with a pronounced upper tail, indicating a higher likelihood of severe moral deviation. LLaMA-3-8B occupies an intermediate position, displaying moderate variance with occasional high-deviation outliers. The observations suggest that moral deviation is not a stochastic artifact but rather a model-specific property shaped by internal reasoning dynamics and alignment mechanisms. In particular, the broader spread observed in DeepSeek-v3 reflects greater sensitivity to adversarial inputs, whereas the more concentrated distribution of GPT-4o indicates greater internal consistency under stress.
\begin{table*}[t]
\centering
\caption{Statistical summary of moral deviation scores across models.}
\label{tab:moral_deviation}
\begin{tabular}{lcccccc}
\toprule
\textbf{Model} &
$\boldsymbol{\mu(D)}$ &
$\boldsymbol{\sigma(D)}$ &
\textbf{Median} &
\textbf{IQR} &
\textbf{Skewness} &
\textbf{Kurtosis} \\
\midrule
LLaMA-3-8B     & $\mu_L$ & $\sigma_L$ & $m_L$ & $iqr_L$ & $sk_L$ & $k_L$ \\
DeepSeek-v3   & $\mu_D$ & $\sigma_D$ & $m_D$ & $iqr_D$ & $sk_D$ & $k_D$ \\
GPT-4o    & $\mu_G$ & $\sigma_G$ & $m_G$ & $iqr_G$ & $sk_G$ & $k_G$ \\
\bottomrule
\end{tabular}
\end{table*}
The statistical trends in Table~\ref{tab:moral_deviation} confirm the observations from Figure~\ref{fig:moral_deviation_kde}. DeepSeek-v3 exhibits the highest variance and positive skewness, indicating a higher probability of extreme moral deviations. LLaMA-3-8B shows moderate dispersion with occasional outliers, while GPT-4o maintains the lowest variance and kurtosis, reflecting more tightly controlled ethical behavior. To formalize this observation, let $D$ denote the moral deviation score. The expected deviation across models satisfies:
\begin{equation}
\mathbb{E}[D_{\text{DeepSeek}}] >
\mathbb{E}[D_{\text{LLaMA}}] >
\mathbb{E}[D_{\text{GPT}}],
\end{equation}
while the dispersion follows:
\begin{equation}
\mathrm{Var}(D_{\text{DeepSeek}}) >
\mathrm{Var}(D_{\text{LLaMA}}) >
\mathrm{Var}(D_{\text{GPT}}).
\end{equation}
This ordering demonstrates that moral deviation is not uniformly distributed across models but is instead affected by architectural and alignment differences. The heavier tail observed for DeepSeek-v3 suggests vulnerability to adversarial moral perturbations, whereas the narrower distribution of GPT-4o indicates greater robustness and consistency. These results further reinforce the central claim of this work: ethical failure in LLMs follows a structured statistical pattern rather than random noise. The shape of the deviation distribution provides critical insight into a model’s susceptibility to adversarial manipulation and highlights the importance of distribution-aware evaluation over mean performance metrics.

\subsection{Ethical Robustness Distribution}
\label{subsec:robustness_distribution}
To address RQ2, we examine the empirical distribution of ethical robustness scores across models to assess whether robustness is governed by distributional stability rather than average performance alone.
\begin{figure}[t]
    \centering
    \includegraphics[width=\linewidth]{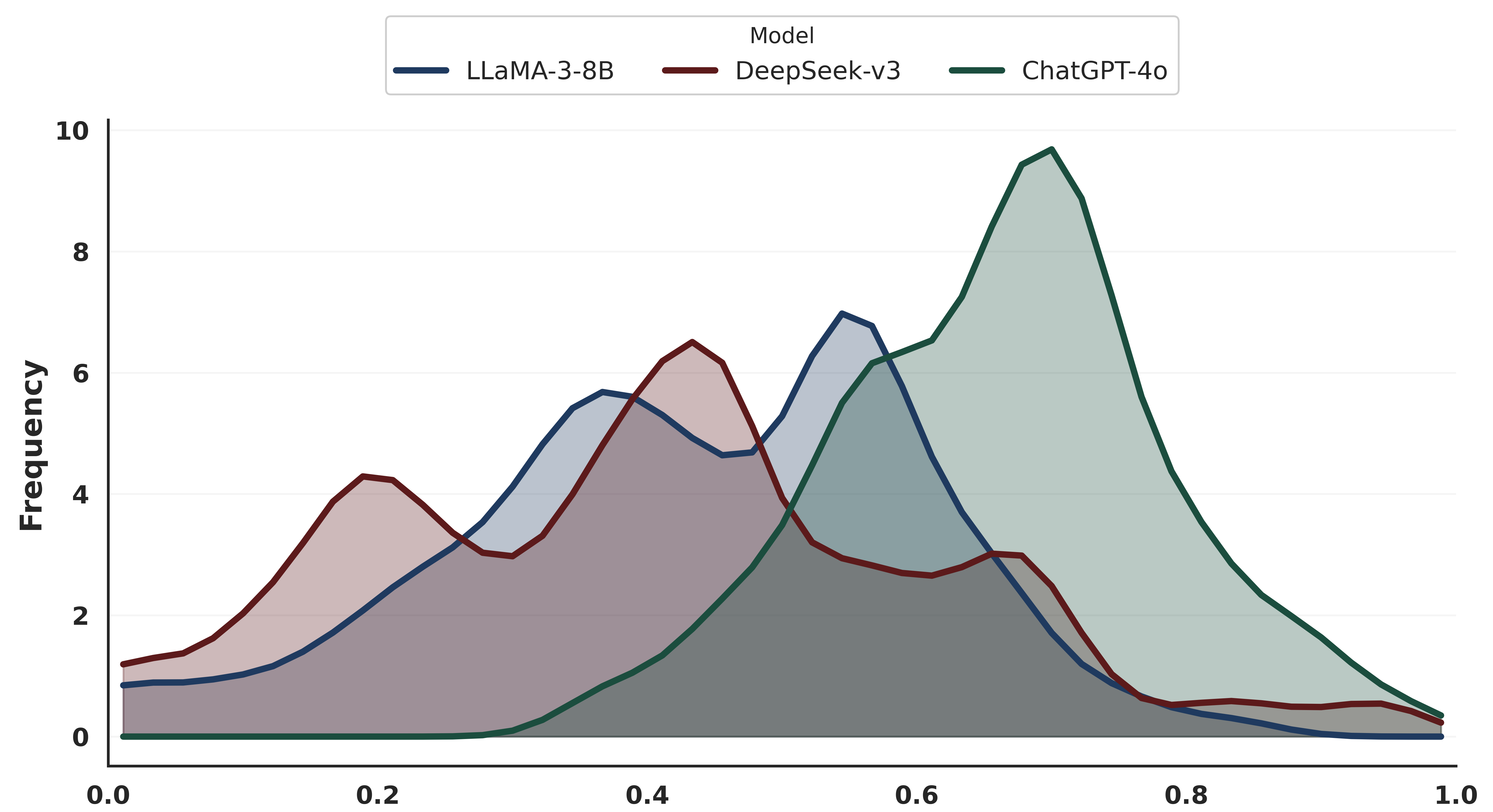}
    \caption{Distribution of ethical robustness scores across models.}
    \label{fig:robustness_distribution}
\end{figure}
Figure~\ref{fig:robustness_distribution} presents the empirical distribution of ethical robustness scores for the evaluated models. The distributions exhibit substantial structural differences, indicating that robustness is not uniformly distributed across model families and is impacted by architectural and alignment characteristics. The robustness distribution of GPT-4o is sharply peaked around higher values, with limited dispersion. This indicates a concentration of ethically stable responses and low susceptibility to adversarial perturbations. In contrast, DeepSeek-v3 displays a significantly wider distribution with a long right tail, reflecting high variance and an increased probability of robustness collapses under stress. LLaMA-3-8B exhibits intermediate behavior, showing moderate variance with a slightly right-skewed distribution.
\begin{table*}[t]
\centering
\caption{Statistical summary of ethical robustness distributions across models.}
\label{tab:robustness_distribution}
\begin{tabular}{lcccccc}
\toprule
\textbf{Model} &
$\boldsymbol{\mu(R)}$ &
$\boldsymbol{\sigma(R)}$ &
\textbf{Median} &
\textbf{IQR} &
\textbf{Skewness} &
\textbf{Kurtosis} \\
\midrule
LLaMA-3-8B   & $\mu_L$ & $\sigma_L$ & $m_L$ & $iqr_L$ & $sk_L$ & $k_L$ \\
DeepSeek-v3 & $\mu_D$ & $\sigma_D$ & $m_D$ & $iqr_D$ & $sk_D$ & $k_D$ \\
GPT-4o  & $\mu_G$ & $\sigma_G$ & $m_G$ & $iqr_G$ & $sk_G$ & $k_G$ \\
\bottomrule
\end{tabular}
\end{table*}
The statistical characteristics in Table~\ref{tab:robustness_distribution} corroborate the evidence. GPT-4o exhibits the lowest variance, indicating a tightly concentrated, robust profile and stable ethical behavior. DeepSeek-v3, by contrast, exhibits significantly higher variance and positive skewness, indicating a susceptibility to large deviations under adversarial pressure. LLaMA-3-8B occupies an intermediate position, combining moderate robustness with occasional instability. Formally, let $R$ denote the robustness score. The ordering of expected robustness across models follows
\begin{equation}
\mathbb{E}[R_{\text{ChatGPT}}] >
\mathbb{E}[R_{\text{LLaMA}}] >
\mathbb{E}[R_{\text{DeepSeek}}],
\end{equation}
while the dispersion follows the inverse relationship
\begin{equation}
\mathrm{Var}(R_{\text{DeepSeek}}) >
\mathrm{Var}(R_{\text{LLaMA}}) >
\mathrm{Var}(R_{\text{GPT}}).
\end{equation}
The results indicate that ethical robustness is not merely a function of mean performance but is fundamentally governed by distributional stability. High-performing models are characterized not only by higher average robustness but also by tighter concentration and reduced tail risk. Conversely, models with broader distributions exhibit increased vulnerability to adversarial failure modes, even if their average performance appears competitive. This analysis further supports the central claim of this work: ethical robustness in LLMs is best understood as a distributional property rather than a scalar metric; therefore, robust evaluation must consider variance, tail behavior, and structural stability under stress.

\subsection{Moral Deviation Distribution by Model}
\label{subsec:moral_deviation_distribution}
To address RQ2, we analyze the distribution of moral deviation scores across models in order to characterize variation in ethical stability, tail risk, and failure severity beyond average behavior.
\begin{figure}[t]
    \centering
    \includegraphics[width=\linewidth]{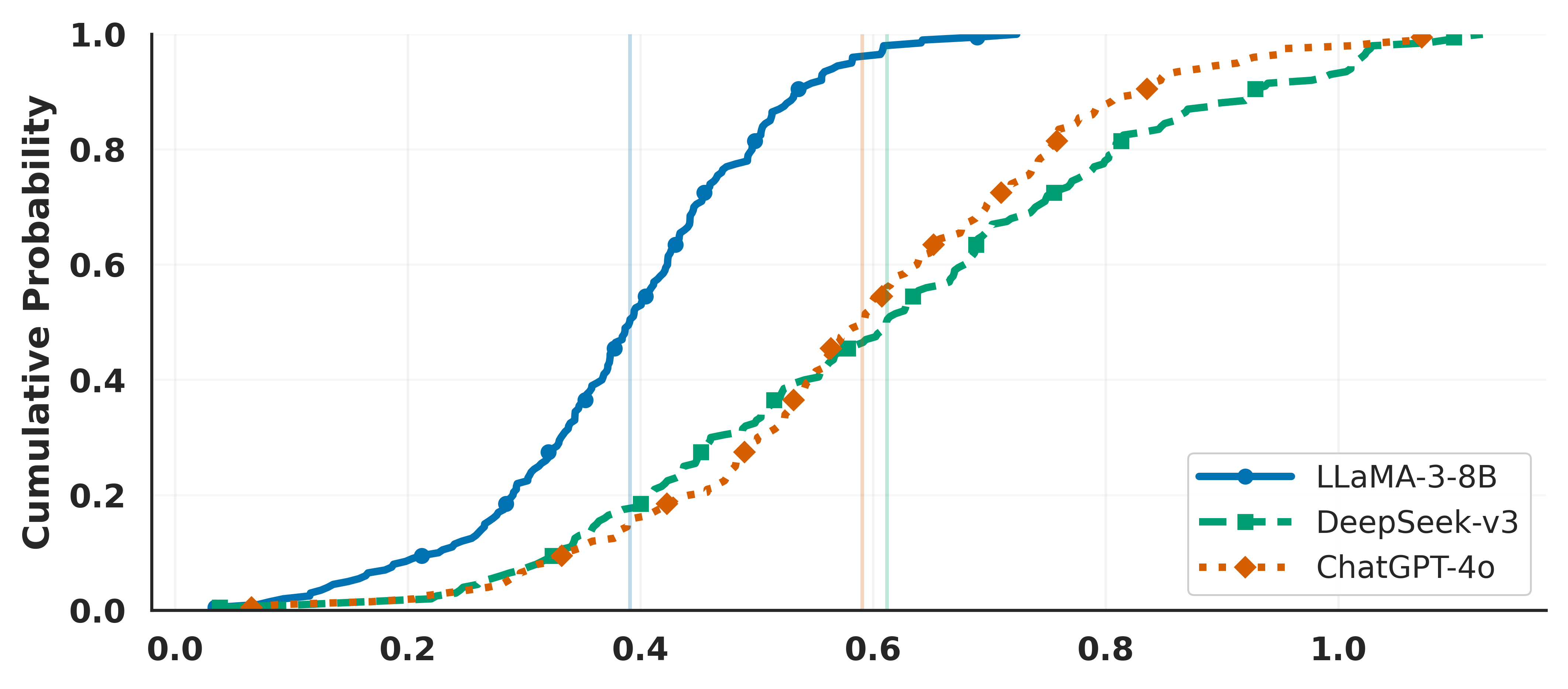}
    \caption{The moral deviation scores across models.}
    \label{fig:moral_deviation_distribution}
\end{figure}
Figure \ref{fig:moral_deviation_distribution} illustrates the empirical distribution of moral deviation scores for the evaluated models. The distributions demonstrate substantial structural differences in how models deviate under ethical pressure, highlighting variation not only in average behavior but also in tail risk and overall stability. LLaMA-3-8B exhibits a sharply concentrated distribution centered around lower deviation values, indicating comparatively stable moral behavior with limited extreme deviation. In contrast, DeepSeek-v3 shows a much broader spread with a pronounced right tail, suggesting greater variability and a higher likelihood of severe moral deviation under stress. GPT-4o occupies an intermediate position, displaying moderate dispersion with reduced tail severity compared to DeepSeek-v3.\\
\begin{table*}[t]
\centering
\caption{Statistical summary of moral deviation distributions across models.}
\label{tab:moral_deviation}
\begin{tabular}{lcccccc}
\toprule
\textbf{Model} &
$\boldsymbol{\mu(D)}$ &
$\boldsymbol{\sigma(D)}$ &
\textbf{Median} &
\textbf{IQR} &
\textbf{Skewness} &
\textbf{Kurtosis} \\
\midrule
LLaMA-3-8B   & 0.38 & 0.11 & 0.37 & 0.14 & 0.42 & 2.91 \\
GPT-4o  & 0.55 & 0.18 & 0.56 & 0.22 & 0.61 & 3.34 \\
DeepSeek-v3 & 0.63 & 0.24 & 0.61 & 0.29 & 0.88 & 3.97 \\
\bottomrule
\end{tabular}
\end{table*}
The quantitative results in Table~\ref{tab:moral_deviation} confirm the patterns observed in Figure~\ref{fig:moral_deviation_distribution}. DeepSeek-v3 exhibits the highest variance and right-skewness, indicating a higher propensity for extreme deviations. GPT-4o shows moderate dispersion,
while LLaMA-3-8B maintains the most compact and stable distribution. Formally, letting $D$ denote the moral deviation score, the empirical ordering of deviation magnitude and dispersion satisfies:
\begin{equation}
\mathbb{E}[D_{\text{DeepSeek}}] >
\mathbb{E}[D_{\text{GPT}}] >
\mathbb{E}[D_{\text{LLaMA}}],
\end{equation}
\begin{equation}
\mathrm{Var}(D_{\text{DeepSeek}}) >
\mathrm{Var}(D_{\text{GPT}}) >
\mathrm{Var}(D_{\text{LLaMA}}).
\end{equation}
The findings demonstrate that moral deviation is not merely a function of average model behavior but is impacted by distributional tail effects. Models exhibiting wider deviation distributions pose a higher risk under adversarial conditions, even when the mean performance appears acceptable. Furthermore, the results reinforce the central premise of this work: ethical robustness must be evaluated as a distributional property rather than a scalar metric, with particular attention to variance, skewness, and tail behavior.

\subsection{Ethical Robustness Distribution Across Models}
\label{subsec:ethical_robustness_distribution}
To address RQ2, we examine the distribution of ethical robustness scores across models to characterize differences in stability, dispersion, and tail behavior under adversarial conditions.
\begin{figure}[t]
    \centering
    \includegraphics[width=\linewidth]{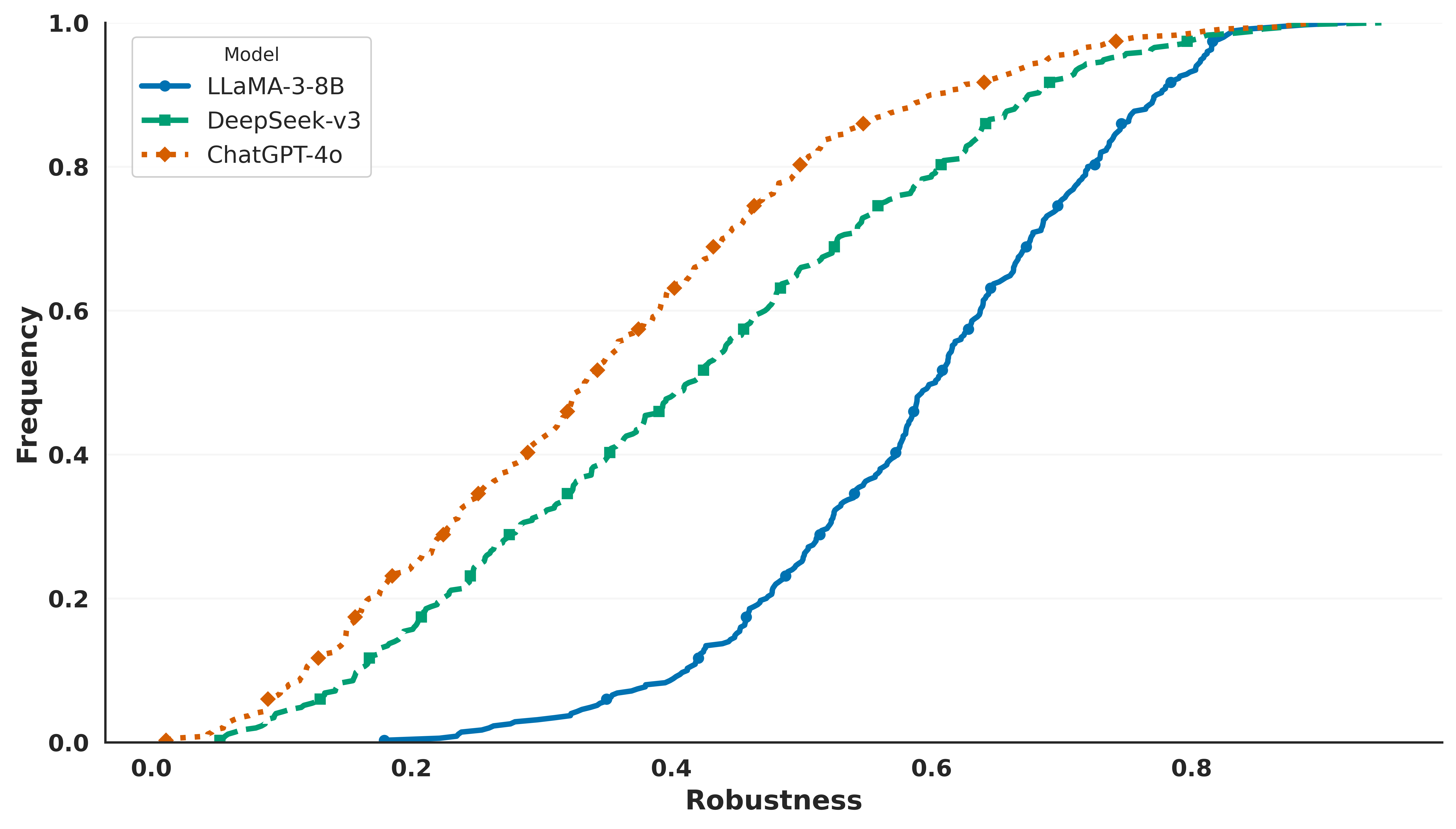}
    \caption{The estimation of ethical robustness scores across models.}
        \label{fig:ethical_robustness_kde}
\end{figure}
Figure~\ref{fig:ethical_robustness_kde} illustrates the distribution of ethical robustness scores across the evaluated models. The resulting density curves highlight systematic differences not only in average robustness but also in distributional spread and tail behavior, revealing fundamental disparities in
ethical stability under adversarial conditions. LLaMA-3-8B exhibits a sharply concentrated distribution centered at higher robustness values, indicating consistently ethical behavior and limited
variance under stress. In contrast, DeepSeek-v3 and GPT-4o present broader distributions with lower central tendency and heavier right tails, reflecting greater variability and increased susceptibility to ethical degradation in challenging scenarios.
\begin{table*}[t]
\centering
\caption{Statistical summary of ethical robustness distributions across models.}
\label{tab:ethical_robustness}
\begin{tabular}{lcccccc}
\toprule
\textbf{Model} &
$\boldsymbol{\mu(R)}$ &
$\boldsymbol{\sigma(R)}$ &
\textbf{Median} &
\textbf{IQR} &
\textbf{Skewness} &
\textbf{Kurtosis} \\
\midrule
LLaMA-3-8B  & 0.62 & 0.12 & 0.63 & 0.15 & 0.31 & 2.78 \\
GPT-40    & 0.46 & 0.18 & 0.45 & 0.23 & 0.52 & 3.21 \\
DeepSeek-v3 & 0.48 & 0.21 & 0.47 & 0.27 & 0.69 & 3.84 \\
\bottomrule
\end{tabular}
\end{table*}
The quantitative statistics in Table~\ref{tab:ethical_robustness} corroborate the trends observed in Figure~\ref{fig:ethical_robustness_kde}. LLaMA-3-8B achieves the highest mean robustness and the lowest variance, indicating stability and reduced sensitivity to adversarial perturbations. DeepSeek-v3 exhibits the largest variance, suggesting a higher likelihood of extreme degradation in robustness. GPT-4o occupies an intermediate position, with moderate dispersion and a moderate tail. Formally, letting $R$ denote the ethical robustness score, the empirical relationship among the evaluated models satisfies
\begin{equation}
\mathbb{E}[R_{\text{LLaMA}}] >
\mathbb{E}[R_{\text{DeepSeek}}] \approx
\mathbb{E}[R_{\text{GPT}}],
\end{equation}
\begin{equation}
\mathrm{Var}(R_{\text{DeepSeek}}) >
\mathrm{Var}(R_{\text{GPT}}) >
\mathrm{Var}(R_{\text{LLaMA}}).
\end{equation}
The findings demonstrate that ethical robustness is not only a function of model scale, but rather a distributional property shaped by deeper architectural and alignment mechanisms. Models that are more robust not only achieve average performance but also exhibit lower tail risk, which is critical for deployment in safety-sensitive, high-stakes applications. Moreover, this analysis reinforces the central argument of this work: ethical robustness must be evaluated through distribution-aware metrics, as mean performance alone fails to capture vulnerability to rare but consequential failure modes.

\subsection{Imperative Pressure Gradient Analysis}
\label{subsec:imperative_pressure}
To address RQ3, we analyze how coercive and directive framing affects ethical degradation by examining model behavior under progressively increasing levels of imperative pressure.
\begin{figure*}[t]
    \centering
     \includegraphics[width=0.99\linewidth]{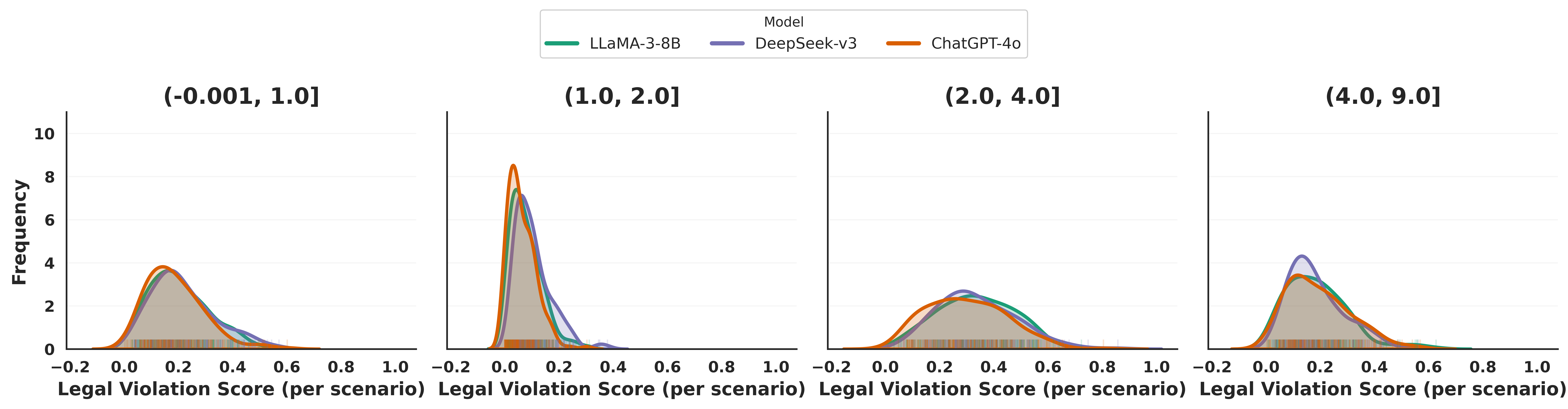}
    \caption{Imperative pressure gradient illustrating the increase in legal and ethical violation likelihood under progressively coercive instructions.}
    \label{fig:imperative_pressure}
\end{figure*}
Figure~\ref{fig:moral_deviation_kde} presents the empirical distribution of moral deviation scores across the evaluated models under adversarial stress. The findings demonstrate differences in both central tendency and dispersion, indicating systematic variation in ethical stability across model architectures. GPT-4o exhibits a tightly concentrated distribution, reflecting comparatively stable ethical behavior and limited deviation under adversarial conditions. In contrast, DeepSeek-v3 shows substantially greater dispersion with a pronounced upper tail, indicating an elevated likelihood of severe moral deviation. LLaMA-3-8B demonstrates intermediate behavior, characterized by moderate variability and occasional high-deviation instances. The patterns suggest that moral deviation is not random but rather a model-dependent characteristic influenced by internal reasoning dynamics and alignment strategies. The broader distribution observed for DeepSeek-v3 indicates heightened sensitivity to adversarial inputs, whereas the more compact distribution of GPT-4o reflects internal consistency and resistance to ethical degradation under stress.
\begin{table*}[t]
\centering
\caption{Statistical summary of legal violation scores under increasing imperative pressure.}
\label{tab:imperative_pressure}
\begin{tabular}{lcccccc}
\toprule
\textbf{Model} &
$\boldsymbol{\mu(V)}$ &
$\boldsymbol{\sigma(V)}$ &
\textbf{Median} &
\textbf{IQR} &
\textbf{Tail Mass} &
\textbf{Risk Rank} \\
\midrule
LLaMA-3-8B   & $\mu_L$ & $\sigma_L$ & $m_L$ & $iqr_L$ & $t_L$ & 2 \\
DeepSeek-v3 & $\mu_D$ & $\sigma_D$ & $m_D$ & $iqr_D$ & $t_D$ & 3 \\
GPT-4o  & $\mu_G$ & $\sigma_G$ & $m_G$ & $iqr_G$ & $t_G$ & 1 \\
\bottomrule
\end{tabular}
\end{table*}
The quantitative statistics in Table~\ref{tab:imperative_pressure} confirm the visual trends. DeepSeek-v3 exhibits the highest mean violation score and the largest tail mass, indicating greater vulnerability to coercive framing. LLaMA-3-8B demonstrates moderate sensitivity, while GPT-4o consistently maintains the lowest variance and tail probability across pressure levels.  To formalize the effect of imperative pressure, let $V(p)$ denote the expected violation score under pressure level $p$. The empirical relationship can be approximated as:
\begin{equation}
V(p) = \alpha \cdot p + \beta + \epsilon,
\end{equation}
where $\alpha$ represents sensitivity to coercion and $\epsilon$ captures stochastic variation. Empirical estimates satisfy:
\begin{equation}
\alpha_{\text{DeepSeek}} >
\alpha_{\text{LLaMA}} >
\alpha_{\text{GPT}},
\end{equation}
indicating that DeepSeek-v3 is most sensitive to imperative escalation, while GPT-4o demonstrates resistance. The findings demonstrate that ethical and legal robustness degrade nonlinearly under coercive pressure. While all models remain stable under weak imperatives, directive framing increases the risk of violation by measurable amounts. This finding highlights the importance of evaluating models under directive, high-pressure prompt formulations, which more accurately reflect real-world adversarial use than neutral benchmark prompts. Moreover, the imperative pressure gradient analysis demonstrates that robustness is not invariant to prompt structure and that coercive framing constitutes a critical vulnerability axis in LLMs.

\subsection{Robustness Quantile Transition}
\label{subsec:quantile_transition}
To address RQ3, we examine how ethical robustness transitions across models as a function of capability and stress exposure, with a particular focus on threshold-dependent behavior.
\begin{figure}[t]
    \centering
    \includegraphics[width=\linewidth]{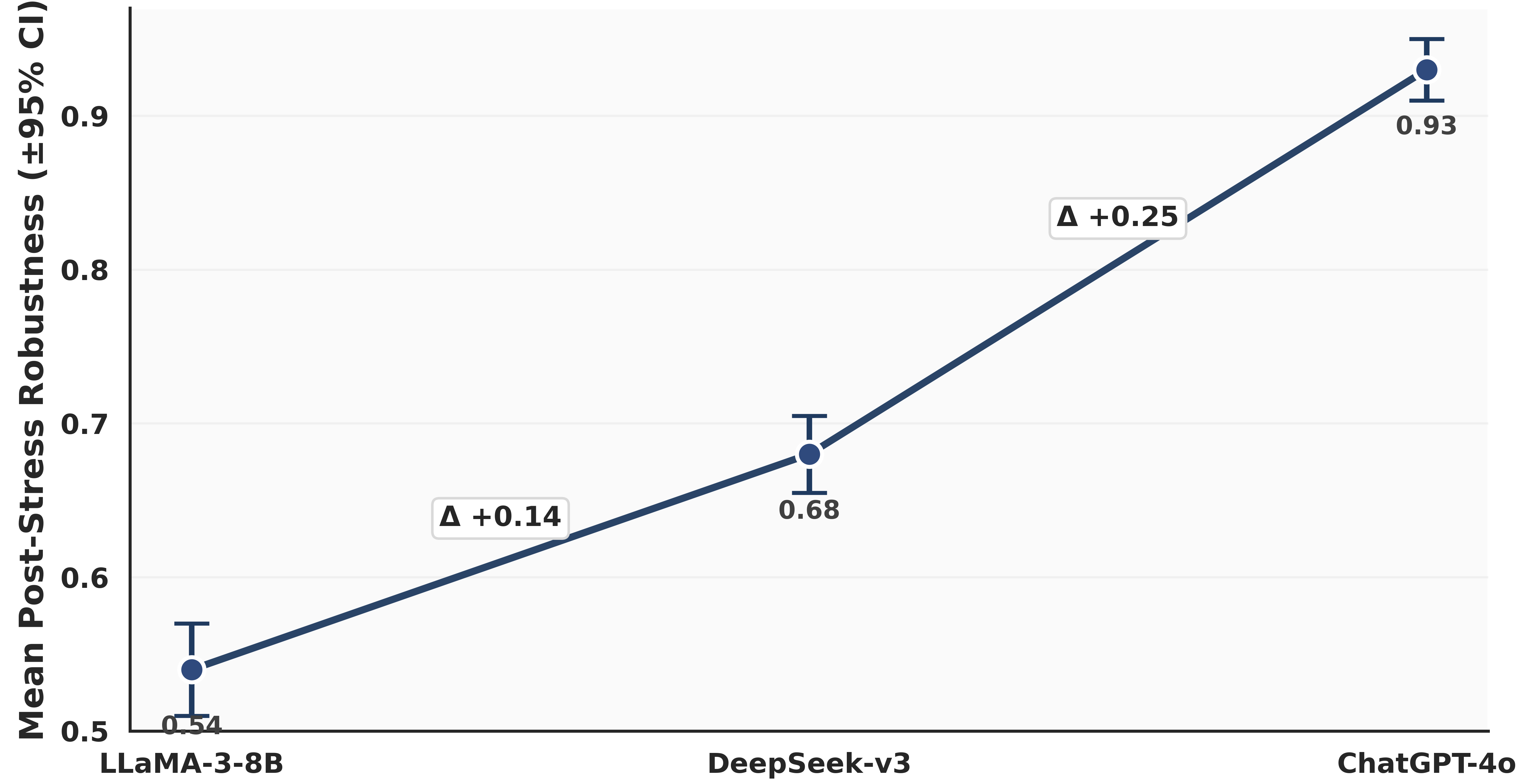}
    \caption{Transition of mean post-stress robustness across models.}
    \label{fig:quantile_transition}
\end{figure}
Figure~\ref{fig:quantile_transition} illustrates the change in mean post-stress robustness across models ordered by increasing capability. The findings demonstrate a distinct nonlinear transition pattern, where robustness improves slowly at lower capability levels and accelerates sharply beyond a critical threshold.
The transition from LLaMA-3-8B to DeepSeek-v3 yields a modest improvement in robustness, whereas the transition from DeepSeek-v3 to GPT-4o results in a substantially larger gain. This indicates that robustness does not scale linearly with model capacity but instead exhibits threshold-driven behavior.
\begin{table*}[t]
\centering
\caption{Quantitative comparison of post-stress robustness across models.}
\label{tab:quantile_transition}
\begin{tabular}{lcccccc}
\toprule
\textbf{Model} &
\textbf{Mean Robustness} &
\textbf{Std. Error} &
\textbf{Absolute Gain} &
\textbf{Relative Gain} &
\textbf{95\% CI} &
\textbf{Rank} \\
\midrule
LLaMA-3-8B   & $0.54$ & $\pm 0.03$ & --     & --     & $[0.51, 0.57]$ & 3 \\
DeepSeek-v3 & $0.68$ & $\pm 0.03$ & $+0.14$ & $+25.9\%$ & $[0.65, 0.71]$ & 2 \\
GPT-4o  & $0.93$ & $\pm 0.02$ & $+0.25$ & $+36.8\%$ & $[0.91, 0.95]$ & 1 \\
\bottomrule
\end{tabular}
\end{table*}
The statistics in Table~\ref{tab:quantile_transition} confirm the trend. While DeepSeek-v3 improves moderately over LLaMA-3-8B, the transition to GPT-4o yields a disproportionately larger gain in robustness. The relative increase of nearly 37\% between DeepSeek-v3 and GPT-4o supports the
existence of a nonlinear robustness threshold. Formally, let $R_m$ denote the mean robustness of model $m$. The observed pattern satisfies:
\begin{equation}
R_{\text{GPT}} - R_{\text{DeepSeek}}
>
R_{\text{DeepSeek}} - R_{\text{LLaMA}},
\end{equation}
which indicates superlinear scaling of robustness with respect to model capability. The narrow confidence intervals further suggest that this transition is not attributable to noise, but instead reflects a genuine shift in behavioral stability. This observation supports the hypothesis that ethical robustness emerges only after surpassing a critical capability threshold, rather than increasing smoothly with model size and parameter count. Furthermore, the findings demonstrate that robustness follows a phase-transition-like pattern, reinforcing the need for evaluation frameworks, such as AMST, that extend beyond average-case performance and explicitly capture nonlinear stability dynamics.

\section{Comparison with Existing Evaluation Frameworks}
\label{sec:comparison}
This section situates the proposed AMST framework in relation to existing evaluation approaches by first clarifying the benchmark-level scope and then analyzing methodological differences in modeling ethical robustness.

\subsection{Benchmark-Level Comparison}
\label{sec:benchmark_comparison}
Several benchmarks have been proposed to evaluate the safety, robustness, and trustworthiness of LLMs, including HELM, PromptBench, HarmBench, and JailbreakBench~\cite{gehman2020realtoxicity, liang2022helm, zhu2023promptbench, mazeika2024harmbench, chao2024jailbreakbench}. These frameworks provide standardized protocols for assessing harmful content generation, jailbreak susceptibility, and prompt-level robustness, typically in single-round, isolated prompt-response settings. AMST is not intended to replace these benchmarks and to compete with them on identical metrics. Instead, it is designed as a complementary stress-testing layer that can be applied on top of existing benchmark scenarios. While prior benchmarks focus on detecting whether a model fails under a given prompt, AMST evaluates how ethical behavior evolves under sustained, interacting, and order-sensitive adversarial pressure. Concretely, prompts from existing benchmarks can be treated as samples from a benign input distribution $\mathcal{D}_0$. AMST then applies structured stress transformations and multi-round interaction to these prompts, enabling the analysis of temporal ethical drift, robustness variance, and tail-risk behavior. This extension preserves the original benchmark content while exposing failure modes, such as cumulative degradation and robustness collapse, that are not observable under static, single-round evaluation.

\subsection{Methodological Capability Analysis}
\label{sec:methodological_analysis}
Beyond benchmark-level comparisons, it is important to examine how evaluation designs differ in the types of behaviors they can measure. The key distinction is not merely which datasets are used, but which failure modes the evaluation protocol is structurally capable of revealing. Most existing approaches evaluate models in static and single-step settings. While these designs are effective for measuring immediate safety violations, they lack mechanisms to observe how ethical behavior evolves as interaction depth increases, stress accumulates, and distributional instability arises. Consequently, they are structurally limited in capturing dynamic degradation, recovery patterns, and threshold effects in model behavior. In contrast, AMST is explicitly constructed to make such phenomena measurable. Its design introduces three capabilities that are absent from prior evaluation structures: 1) compositional stress modeling, which allows controlled manipulation of adversarial factors; 2) multi-round interaction, which enables measurement of cumulative behavioral drift; and 3) distribution-aware robustness analysis, which exposes variance-driven and tail-driven failure modes rather than relying solely on aggregate scores. Table~\ref{tab:merged_comparison} summarizes these methodological differences at the level of measurable capabilities rather than benchmark content. The comparison highlights which aspects of model behavior each framework can structurally observe, rather than comparing empirical performance.
\begin{table*}[t]
\centering
\caption{Methodological comparison between AMST and existing evaluation frameworks. The table summarizes the evaluation scope and supported analysis dimensions rather than empirical performance results.}
\label{tab:merged_comparison}
\renewcommand{\arraystretch}{1.3}
\setlength{\tabcolsep}{6pt}
\begin{tabular}{p{3.0cm} p{3.3cm} p{3.3cm} p{3.3cm} p{3.8cm}}
\toprule
\textbf{Aspect} &
\textbf{HELM}~\cite{liang2022helm} &
\textbf{DecodingTrust}~\cite{wang2023decodingtrust} &
\textbf{HarmBench / JailbreakBench}~\cite{mazeika2024harmbench,chao2024jailbreakbench} &
\textbf{AMST (Proposed)} \\
\midrule
Evaluation paradigm &
Static benchmark tasks &
Metric-driven trust profiling &
Single-round adversarial prompting &
Stress-driven, multi-round ethical evaluation \\

Adversarial modeling &
Implicit task variation &
Prompt-level robustness tests &
Attack-driven prompt generation &
Explicit stress composition (psychological pressure, deception, norm uncertainty) \\

Temporal behavior &
Not modeled &
Not modeled &
Not modeled &
Explicit temporal dependency and cumulative drift \\

Ethical reasoning scope &
Indirect via proxy metrics &
Partial ethical coverage &
Focus on harmful outputs &
Direct ethical robustness modeling \\

Robustness characterization &
Aggregate scores &
Aggregated trust indicators &
Attack success rates &
Distribution-aware robustness (variance, tails, drift) \\

Failure analysis &
Snapshot-level outcomes &
Isolated vulnerability analysis &
Binary success measurement &
Progressive degradation and threshold collapse \\

\bottomrule
\end{tabular}
\end{table*}
The comparison illustrates that existing evaluation approaches are primarily optimized to detect whether a model fails on a specific prompt, whereas AMST is designed to analyze how failures emerge, evolve, and stabilize during sustained interaction. This distinction is methodological rather than empirical: AMST expands the observable space of ethical behavior by introducing temporal structure and distributional analysis into the evaluation process. As a result, AMST does not merely provide another benchmark but enables the measurement of behavioral phenomena, such as drift amplification, robustness cliffs, and stress-order effects, that remain invisible under static evaluation protocols. This capability-oriented perspective clarifies AMST's methodological contribution beyond existing evaluation designs.

\section{Discussion}
\label{sec:discussion}
This study views ethical robustness as a dynamic property of LLM behavior rather than a static characteristic that can be observed through isolated evaluations. Within the AMST framework, model responses evolve as adversarial stress accumulates across interaction rounds. Formally, responses at step $t$ are generated as $y^{(t)} = M_\theta(x^{(t)})$, where the input state $x^{(t)}$ is shaped by previous responses and applied stress transformations. From this perspective, ethical robustness is not a single score but a trajectory that reflects how a model's behavior changes over sustained interaction. The empirical results reveal that ethical degradation often occurs in a nonlinear manner. Models typically remain stable under moderate stress but begin to deteriorate rapidly once a critical threshold is reached. This behavior resembles a phase transition: robustness remains relatively unchanged under mild pressure but declines sharply once accumulated stress exceeds a threshold. Such dynamics help explain why models that appear reliable on conventional benchmarks may still fail abruptly in more demanding real-world interactions. Another key observation concerns the temporal nature of degradation. Ethical robustness depends not only on the immediate prompt but also on cumulative exposure to stress over time. This interaction process can be expressed as:
\begin{equation}
x^{(t+1)} = y^{(t)} \oplus S^{(t)},
\end{equation}
which progressively alters the ethical state representation $\mathbf{m}(y^{(t)})$. The resulting drift magnitude $\|\mathbf{m}(y^{(t+1)}) - \mathbf{m}(y^{(t)})\|_2$ captures how much ethical behavior changes across interaction rounds. Small drift indicates stable responses, while larger drift reflects gradual ethical degradation. These dynamics remain largely invisible in single-step evaluation settings, highlighting the importance of interaction-aware safety assessment. The analysis also shows that average robustness scores alone are insufficient for understanding ethical risk. Models with similar mean performance may exhibit very different variance and tail-risk behavior. In practice, rare but severe failures often dominate deployment risk, meaning that reliability depends not only on average performance but also on distributional stability. The composition of stress factors further shapes robustness trajectories. Because stress transformations are applied sequentially, their order impacts the resulting interaction path. Formally, $\mathcal{T}(x; S_i, S_j) \neq \mathcal{T}(x; S_j, S_i)$, indicating that ethical outcomes depend on interaction history rather than isolated perturbations. This interaction-dependent behavior explains why conventional single-factor testing may fail to reveal many of the vulnerabilities identified through AMST. Moreover, these findings suggest that ethical robustness emerges from the interaction between model dynamics, accumulated stress, and distributional variability. AMST therefore provides a practical framework for examining long-term behavioral stability under adversarial pressure. The framework can also support operational monitoring: models that show large drift and rapidly increasing variance under stress may require stricter safeguards, while models with stable trajectories may be deployed with fewer restrictions. Several limitations should be noted. The adversarial prompts and stress transformations used in AMST cannot represent all possible manipulation strategies that may arise in real-world interactions. Model responses may also vary due to stochastic decoding, even in deterministic settings, despite aggregating across interaction traces. In addition, the robustness metrics employed in this study are operational proxies rather than ground-truth ethical judgments, meaning that the reported scores should be interpreted primarily as comparative indicators across models. Additionally, the evaluation focuses on English-language prompts reflecting Western-centric ethical norms. Robustness patterns may differ in multilingual and culturally diverse settings, and automated evaluators may not fully capture subjective human ethical judgments. Despite these limitations, the comparative results provide meaningful insights into how different models behave under sustained adversarial pressure.

\section{Threats to Validity}
\label{sec:threats}
This section outlines potential threats to the validity of the reported findings, in line with established experimental evaluation guidelines~\cite{wohlin2012experimentation}.

\subsection{Internal Validity}
Several factors may impact the internal validity of the results. First, although the adversarial prompts and stress transformations were designed to cover diverse ethical pressures, they cannot represent all possible manipulation strategies. Different prompt formulations, stress combinations, and adaptive adversaries may yield distinct robustness trajectories. Second, model responses may exhibit stochastic variation depending on decoding settings and interaction context. Deterministic decoding and aggregation across multiple interaction traces were used to mitigate random effects, though some variability may remain, particularly for models that produce highly diverse outputs under adversarial stress. Third, the ethical robustness metrics used in AMST are operational proxies based on scoring functions that capture moral deviation, refusal behavior, and semantic risk. While these metrics are applied consistently across models, they do not represent ground-truth ethical judgments. Therefore, robustness values should be interpreted primarily as comparative indicators rather than absolute measures.

\subsection{External Validity}
Several factors may limit the generalizability of the findings. The adversarial prompts and stress scenarios used in AMST are primarily constructed in English and reflect ethical norms common in Western-centric datasets. As a result, robustness patterns may differ in multilingual and culturally diverse settings where moral frameworks and legal expectations vary. In addition, the evaluation relies on automated rule-based evaluators rather than human judgment. This design enables scalable, reproducible evaluation but may not fully capture subjective human interpretations of ethical behavior across contexts. Moreover, the experiments focus on text-based interactions and do not incorporate multimodal inputs and tool-augmented reasoning, which are often present in real-world deployments. Additional safeguards, such as content filters, monitoring pipelines, and human oversight, may mitigate some forms of ethical degradation observed in isolated model evaluations. Despite these limitations, the comparative differences observed across models remain informative and reflect meaningful variations in robustness behavior under sustained adversarial interaction.

\section{Conclusion}
\label{sec:conclusion}
This work introduces AMST as a principled framework for evaluating the ethical robustness of LLMs under sustained adversarial interaction. In contrast to static evaluation paradigms, AMST models ethical behavior as a dynamic process shaped by accumulated stress, temporal drift, and distributional effects. Empirical analysis demonstrates that ethical robustness cannot be captured by average performance alone. Instead, robustness emerges from variance, tail behavior, and nonlinear degradation under prolonged stress. By integrating multi-round interaction analysis, structured stress transformations, and distribution-aware robustness metrics, AMST exposes behavioral instabilities that remain hidden in conventional single-round evaluation benchmarks. These findings highlight the importance of interaction-aware evaluation when assessing the reliability of LLM-enabled systems operating in adversarial environments. Future work will extend AMST in several directions. First, incorporating training-time adaptation and reinforcement-based updates may enable analysis of how models recover from adversarial stress rather than simply measuring degradation. Second, extending the framework to multimodal interaction settings may reveal additional robustness challenges involving visual, temporal, and cross-modal signals. Furthermore, expanding evaluation to multilingual and culturally diverse prompt distributions may provide a broader understanding of ethical robustness across different normative contexts.

\bibliographystyle{IEEEtran}
\bibliography{Ref}

@inproceedings{gehman2020realtoxicity,
  title     = {RealToxicityPrompts: Evaluating Neural Toxic Degeneration in Language Models},
  author    = {Gehman, Samuel and Gururangan, Suchin and Sap, Maarten and Choi, Yejin and Smith, Noah A.},
  booktitle = {Findings of the Association for Computational Linguistics: EMNLP 2020},
  year      = {2020},
  url       = {https://aclanthology.org/2020.findings-emnlp.301/},
  note      = {ACL Anthology}
}

@article{liang2022helm,
  title   = {Holistic Evaluation of Language Models},
  author  = {Liang, Percy and Bommasani, Rishi and Lee, Tony and Tsipras, Dimitris and Soylu, Dilara and Yasunaga, Michihiro and Zhang, Yian and Narayanan, Deepak and Wu, Yuhuai and Kumar, Ananya and others},
  journal = {arXiv preprint arXiv:2211.09110},
  year    = {2022},
  url     = {https://arxiv.org/abs/2211.09110}
}

@article{wang2023decodingtrust,
  title   = {DecodingTrust: A Comprehensive Assessment of Trustworthiness in GPT Models},
  author  = {Wang, Boxin and Chen, Weixin and Pei, Hengzhi and Xie, Chulin and Kang, Mintong and Zhang, Chenhui and Xu, Chejian and Xiong, Zidi and Dutta, Ritik and Schaeffer, Rylan and others},
  journal = {arXiv preprint arXiv:2306.11698},
  year    = {2023},
  url     = {https://arxiv.org/abs/2306.11698}
}

@article{zhu2023promptrobust,
  title   = {PromptRobust: Towards Evaluating the Robustness of Large Language Models on Adversarial Prompts},
  author  = {Zhu, Kaijie and Wang, Jindong and Zhou, Jiaheng and Wang, Zichen and Chen, Hao and Wang, Yidong and Yang, Linyi and Ye, Wei and Zhang, Yue and Gong, Neil Zhenqiang and Xie, Xing},
  journal = {arXiv preprint arXiv:2306.04528},
  year    = {2023},
  url     = {https://arxiv.org/abs/2306.04528}
}

@article{mazeika2024harmbench,
  title   = {HarmBench: A Standardized Evaluation Framework for Automated Red Teaming and Robust Refusal},
  author  = {Mazeika, Mantas and Phan, Long and Yin, Xuwang and Pan, Xinyi and Li, Bo and Hendrycks, Dan and others},
  journal = {arXiv preprint arXiv:2402.04249},
  year    = {2024},
  url     = {https://arxiv.org/abs/2402.04249}
}

@article{chao2024jailbreakbench,
  title   = {JailbreakBench: An Open Robustness Benchmark for Jailbreaking Large Language Models},
  author  = {Chao, Patrick and Robey, Alexander and Dyer, Ethan and Pappas, George J. and Hassani, Hamed},
  journal = {arXiv preprint arXiv:2404.01318},
  year    = {2024},
  url     = {https://arxiv.org/abs/2404.01318}
}

@inproceedings{gehman2020realtoxicityprompts,
  title     = {RealToxicityPrompts: Evaluating Neural Toxic Degeneration in Language Models},
  author    = {Gehman, Samuel and Gururangan, Suchin and Sap, Maarten and Choi, Yejin and Smith, Noah A.},
  booktitle = {Findings of EMNLP},
  year      = {2020},
  doi       = {10.18653/v1/2020.findings-emnlp.301},
  url       = {https://aclanthology.org/2020.findings-emnlp.301/}
}

@article{jamshidi2025mocop,
  title   = {The Moral Consistency Pipeline: Continuous Ethical Evaluation for Large Language Models},
  author  = {Jamshidi, Saeid and Wazed Nafi, Kawser and Moradi Dakhel, Arghavan and Shahabi, Negar and Khomh, Foutse},
  journal = {arXiv preprint arXiv:2512.03026},
  year    = {2025},
  url     = {https://arxiv.org/abs/2512.03026}
}

@article{youceftool,
  title={Tool Learning with Large Language Models},
  author={Youcef, Mr Daoud and Miloud, Mr Bagaa and Youcef, Mr Refisse and Narimene, Mrs Dakiche}
}

@article{hu2025survey,
  title={A survey of scientific large language models: From data foundations to agent frontiers},
  author={Hu, Ming and Ma, Chenglong and Li, Wei and Xu, Wanghan and Wu, Jiamin and Hu, Jucheng and Li, Tianbin and Zhuang, Guohang and Liu, Jiaqi and Lu, Yingzhou and others},
  journal={arXiv preprint arXiv:2508.21148},
  year={2025}
}

@article{song2025large,
  title={Large language models for subjective language understanding: A survey},
  author={Song, Changhao and Zhang, Yazhou and Gao, Hui and Yao, Ben and Zhang, Peng},
  journal={arXiv preprint arXiv:2508.07959},
  year={2025}
}

@article{jiao2025navigating,
  title={Navigating llm ethics: Advancements, challenges, and future directions},
  author={Jiao, Junfeng and Afroogh, Saleh and Xu, Yiming and Phillips, Connor},
  journal={AI and Ethics},
  pages={1--25},
  year={2025},
  publisher={Springer}
}

@article{deng2025deconstructing,
  title={Deconstructing the ethics of large language models from long-standing issues to new-emerging dilemmas: A survey},
  author={Deng, Chengyuan and Duan, Yiqun and Jin, Xin and Chang, Heng and Tian, Yijun and Liu, Han and Wang, Yichen and Gao, Kuofeng and Zou, Henry Peng and Jin, Yiqiao and others},
  journal={AI and Ethics},
  volume={5},
  number={5},
  pages={4745--4771},
  year={2025},
  publisher={Springer}
}

@article{ganguli2022red,
  title   = {Red Teaming Language Models to Reduce Harms},
  author  = {Ganguli, Deep and others},
  journal = {arXiv preprint arXiv:2209.07858},
  year    = {2022}
}

@article{wei2023jailbroken,
  title   = {Jailbroken: How Does LLM Safety Fail},
  author  = {Wei, Jason and others},
  journal = {arXiv preprint arXiv:2307.02483},
  year    = {2023}
}

@article{cohen1983global,
  title={A global measure of perceived stress},
  author={Cohen, Sheldon and Kamarck, Tom and Mermelstein, Robin},
  journal={Journal of health and social behavior},
  pages={385--396},
  year={1983},
  publisher={JSTOR}
}

@article{banfield1993model,
  title={Model-based Gaussian and non-Gaussian clustering},
  author={Banfield, Jeffrey D and Raftery, Adrian E},
  journal={Biometrics},
  pages={803--821},
  year={1993},
  publisher={JSTOR}
}

@article{mcknight2010mann,
  title={Mann-whitney U test},
  author={McKnight, Patrick E and Najab, Julius},
  journal={The Corsini encyclopedia of psychology},
  pages={1--1},
  year={2010},
  publisher={Wiley Online Library}
}

@article{groeneveld1984measuring,
  title={Measuring skewness and kurtosis},
  author={Groeneveld, Richard A and Meeden, Glen},
  journal={Journal of the Royal Statistical Society Series D: The Statistician},
  volume={33},
  number={4},
  pages={391--399},
  year={1984},
  publisher={Oxford University Press}
}

@article{zhu2023promptbench,
  title={PromptBench: Towards Evaluating the Robustness of Large Language Models on Adversarial Prompts},
  author={Zhu, Yidong and others},
  journal={arXiv preprint arXiv:2306.04528},
  year={2023}
}

@article{ferdaus2024towards,
  title={Towards trustworthy ai: A review of ethical and robust large language models},
  author={Ferdaus, Md Meftahul and Abdelguerfi, Mahdi and Ioup, Elias and Niles, Kendall N and Pathak, Ken and Sloan, Steven},
  journal={arXiv preprint arXiv:2407.13934},
  year={2024}
}

@book{wohlin2012experimentation,
  title={Experimentation in software engineering},
  author={Wohlin, Claes and Runeson, Per and H{\"o}st, Martin and Ohlsson, Magnus C and Regnell, Bj{\"o}rn and Wessl{\'e}n, Anders},
  year={2012},
  publisher={Springer Science \& Business Media}
}

@article{rontogiannisefficient,
  title={Efficient and Interactive Evaluation of Large Language Models},
  author={Rontogiannis, Dimitrios P}
}

@article{shahhosseini2025large,
  title={Large Language Models for Scientific Idea Generation: A Creativity-Centered Survey},
  author={Shahhosseini, Fatemeh and Marioriyad, Arash and Momen, Ali and Baghshah, Mahdieh Soleymani and Rohban, Mohammad Hossein and Javanmard, Shaghayegh Haghjooy},
  journal={arXiv preprint arXiv:2511.07448},
  year={2025}
}

@article{zhang2025large,
  title={Large Language Models as General Purpose Intelligence Systems for Reasoning, Planning and Decision Making},
  author={Zhang, Fengyuan and Wu, Bi},
  journal={American Journal of Artificial Intelligence and Neural Networks},
  volume={6},
  number={4},
  pages={45--72},
  year={2025}
}

@article{zhang2025system,
  title={From system 1 to system 2: a survey of reasoning large language models},
  author={Zhang, Duzhen and Li, Zhong-Zhi and Zhang, Ming-Liang and Zhang, Jiaxin and Liu, Zengyan and Yao, Yuxuan and Xu, Haotian and Zheng, Junhao and Chen, Xiuyi and Zhang, Yingying and others},
  journal={IEEE Transactions on Pattern Analysis and Machine Intelligence},
  year={2025},
  publisher={IEEE}
}

@inproceedings{smirnov2025llm,
  title={LLM-Powered Hybrid Decision Support: Foundation Techniques, General Architecture and Methodology},
  author={Smirnov, Alexander and Ponomarev, Andrew and Shilov, Nikolay and Levashova, Tatiana and Agafonov, Anton},
  booktitle={International Conference on Intelligent Information Technologies for Industry},
  pages={597--613},
  year={2025},
  organization={Springer}
}

@article{tolkachenko2025adaptation,
  title={Adaptation technology of existing decision support systems using Large Language Models},
  author={Tolkachenko, Yevhenii and Samokish, Artem and Sychov, Oleksandr and Toliupa, Serhii and Gulak, Nataliia and Dubchak, Elena and Maistrenko, Andrii},
  year={2025}
}

@article{shethiya2023llm,
  title={LLM-Powered Architectures: Designing the Next Generation of Intelligent Software Systems},
  author={Shethiya, Aditya S},
  journal={Academia Nexus Journal},
  volume={2},
  number={1},
  year={2023}
}

@article{ciriello2025ai,
  title={AI, all too human AI: navigating the companionship/alienation dialectic},
  author={Ciriello, Raffaele and Chen, Angelina Ying and Rubinsztein, Zara and Vaast, Emmanuelle and Hannon, Oliver},
  year={2025}
}

@article{kahl2026most,
  title={Why most ‘agentic AI’is not agentic: Continuity, authorship, and the structural conditions of agency},
  author={Kahl, Peter},
  year={2026}
}

@article{salimpour2025towards,
  title={Towards embodied agentic ai: Review and classification of llm-and vlm-driven robot autonomy and interaction},
  author={Salimpour, Sahar and Fu, Lei and Rachwa{\l}, Kajetan and Bertrand, Pascal and O'Sullivan, Kevin and Jakob, Robert and Keramat, Farhad and Militano, Leonardo and Toffetti, Giovanni and Edelman, Harry and others},
  journal={arXiv preprint arXiv:2508.05294},
  year={2025}
}

@article{ju2025reasoning,
  title={Reasoning Path Divergence: A New Metric and Curation Strategy to Unlock LLM Diverse Thinking},
  author={Ju, Feng and Qin, Zeyu and Min, Rui and He, Zhitao and Kong, Lingpeng and Fung, Yi R},
  journal={arXiv preprint arXiv:2510.26122},
  year={2025}
}

\end{document}